\documentclass[10pt,journal, twocolumn]{IEEEtran}
\ifCLASSOPTIONcompsoc
\else
\fi

\ifCLASSINFOpdf
\else
\fi

\usepackage[table]{xcolor}
\usepackage{times}
\usepackage{epsfig}
\usepackage{graphicx}
\usepackage{amsmath}
\usepackage{amssymb}
\usepackage{epsfig}
\usepackage{graphicx}
\usepackage{amsmath}
\usepackage{amssymb}
\usepackage{bbm}

\usepackage{array,setspace, amsfonts,url,bm, cellspace}
\usepackage{tikz}
\usepackage{epstopdf}
\usetikzlibrary{positioning}
\usetikzlibrary{arrows}
\usepackage{pifont}
\usepackage{multirow, hhline}

\newcolumntype{M}[1]{>{\centering\arraybackslash}m{#1}}

\usepackage[pagebackref=true,breaklinks=true,letterpaper=true,colorlinks,bookmarks=false]{hyperref}

\newcommand{\etal}{\textit{et al}.}

\begin{document}

\title{A Symbolic Temporal Pooling method for Video-based
Person Re-Identification}

\author{S V Aruna Kumar, Ehsan Yaghoubi, Hugo Proen\c{c}a,~\IEEEmembership{Senior Member,~IEEE}
\IEEEcompsocitemizethanks{Authors are with the IT: Instituto de Telecomunica\c{c}\~{o}es, Department of Computer Science, University of Beira Interior, Covilh\~{a}, Portugal, E-mail: arunkumarsv55@gmail.com, ehsan.yaghoubi@gmail.com, hugomcp@di.ubi.pt }
\thanks{Manuscript received: ?, 2020}}


\IEEEtitleabstractindextext{%
\begin{abstract}
In video-based person re-identification, both the spatial and temporal features are known to provide orthogonal cues to effective representations. Such representations are currently typically obtained by aggregating the frame-level features using \textit{max/avg} pooling, at different points of the models. However, such operations also decrease the amount of discriminating information available, which is particularly hazardous in case of poor separability between the different classes. To alleviate this problem, this paper introduces a symbolic temporal pooling method, where frame-level features are represented in the distribution valued symbolic form, yielding from fitting an Empirical Cumulative Distribution Function (ECDF) to each feature. Also, considering that the original triplet loss formulation cannot be applied directly to this kind of representations, we introduce a symbolic triplet loss function that infers the similarity between two symbolic objects. Having carried out an extensive empirical evaluation of the proposed solution against the state-of-the-art, in four well known data sets (MARS, iLIDS-VID, PRID2011 and P-DESTRE), the observed results point for consistent improvements in performance over the previous best performing techniques. Under a reproducible research paradigm, both the code and the empirical validation framework are available at \url{https://github.com/aru05c/SymbolicTemporalPooling}.
\end{abstract}

}

\maketitle

\IEEEdisplaynontitleabstractindextext

\IEEEpeerreviewmaketitle



    

    
    
    

\section{Introduction}

\IEEEPARstart{P}erson Re-Identification (re-id) refers to the cross-camera retrieval task, in which a query from a target subject is used to retrieve identities in a gallery set. Many obstacles can be posed to this process, such as variations in human poses, lighting conditions, partial occlusions and cluttered backgrounds. Person re-id methods can be broadly categorized into two types: image- (e.g.,\cite{liu2017end,zhao2017deeply,wang2018mancs}) and video-based (e.g.,  \cite{gu2019temporal,liu2019,rao2018video}). Recently, more attention has been given to video-based re-id, as it fuses information from both the spatial and temporal domains, which alleviates the difficulties in case of poor separability problems, such as the visual surveillance in outdoor environments. In this context, most of the existing methods use frame-level feature extractors, and then exploit the temporal cues according to some aggregation function. 

The strategies for extracting such kind of representations can be divided into three categories: i) optical-flow based; ii) 3D-CNN based; and iii) using temporal aggregation techniques. In the first category of methods, the optical flow describes the dynamic features that feed the CNN inputs  \cite{simonyan2014two,feichtenhofer2017spatiotemporal}. In 3D-CNN based methods, spatial-temporal features are extracted directly from video sequences, according to 3D-CNN models \cite{tran2015learning,qiu2017learning}. In the last category of methods, either Recurrent Neural Networks (RNN)\cite{mclaughlin2016recurrent,liu2019spatial} or temporal pooling \cite{zheng2016mars,li2018diversity} techniques aggregate the frame-level features. 

In particular, when considering the temporal pooling family of methods (where our method belongs to), the  \textit{max} or \textit{avg} pooling functions are the most frequently to aggregate the temporal features, and consider all frames in a sequence equally important. The primary hypothesis in this paper is that, by exhibiting different kinds of singularities and levels of correlation with others, frames in a sequence are not equally important and should be considered in an specific way by any robust classification model. Also, considering that the existing \emph{crisp} data representations fail to capture such kind of variations, the key novelty is to use symbolic representation techniques for such purpose. Hence, in this paper we describe an integrated deep-learning classification framework based in symbolic data representations, that is composed of a feature aggregation module and a symbolic loss function.

Symbolic data representations comprise a myriad of techniques, either based in interval, distribution, periodic or histogram features (\cite{Diday88,alaei2017efficient,matsui2014comparison,verde2015dimension}). Inspired by the successful history of symbolic data analysis (e.g., \cite{guru2010symbolic, harish2010symbolic, de2007fuzzy, IEEE-TCASII00:IAVQ,alaei2017efficient}), we propose a method to capture the variations among the frames in a sequence, while keeping the  discriminating information that is partially disregarded by temporal aggregation functions. 

\begin{figure*}
   \includegraphics[width=\textwidth,,scale=1]{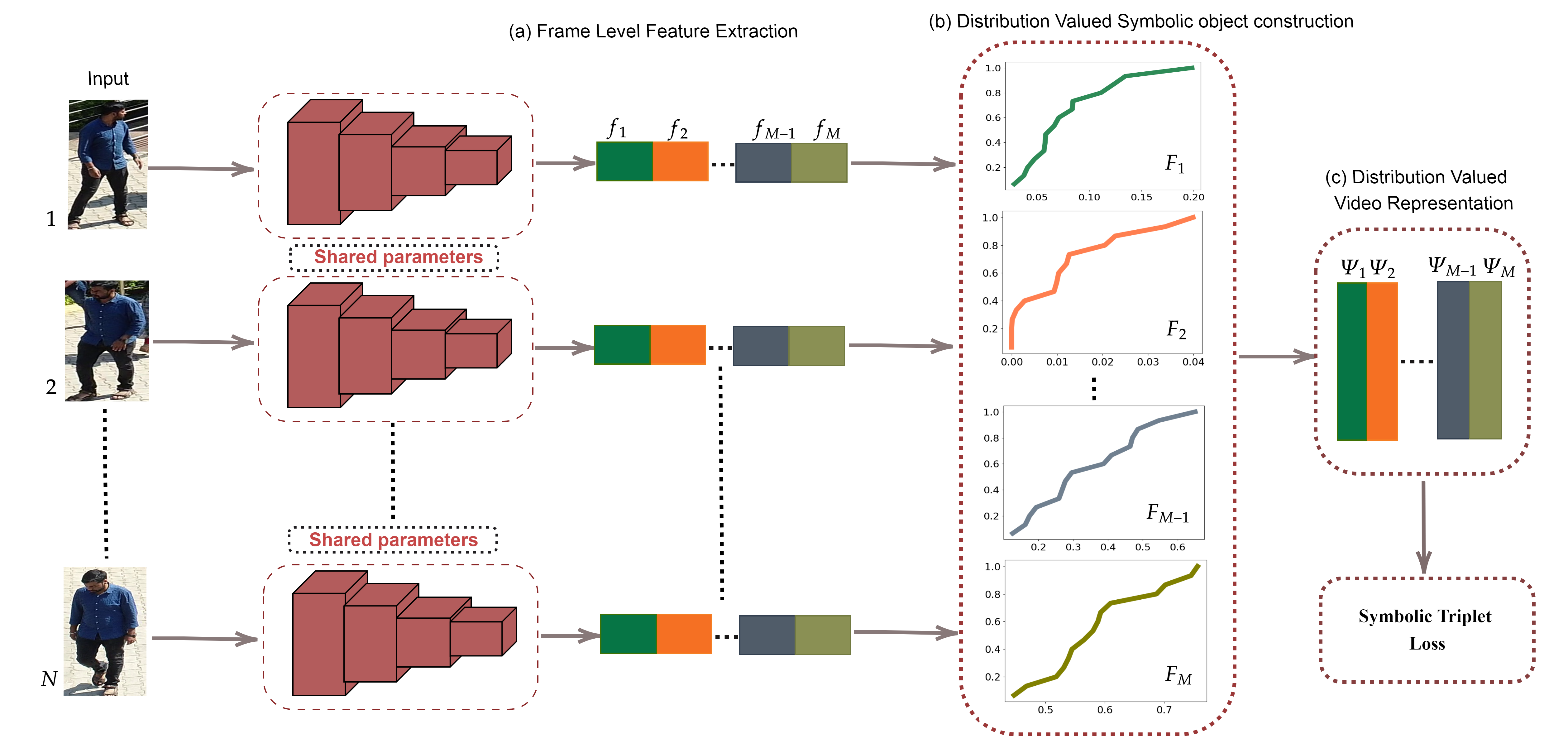}
   \caption{Overview of the proposed model. a) $N$ frames feed a feature extractor model,to obtain $M$ frame-level features (each color represents an individual feature); b) a distribution-valued symbolic object is obtained for each feature by fitting an empirical Cumulative Distribution Function (ECDF) to each one; and c) the values from each  distributional function are concatenated and constitute the symbolic representation used in learning.}
   \label{overview}
 \end{figure*}

Fig. \ref{overview} provides an overview of the processing chain, composed of three steps: i) frame level feature extraction; ii) symbolic data representation;  and iii) learning, according to a symbolic loss function. Initially, frame-level features are extracted using a well known backbone architecture. Next, we convert such \emph{crisp} features into their symbolic representation, by fitting a distributional function for each one. In this setting, each function expresses the variability of a feature across the frames of each sequence. Finally, using the proposed symbolic triplet loss function, we learn a  classification model. 

Having conducted an extensive experimental validation in four publicly available video-based re-id datasets( MARS \cite{zheng2016mars}, iLIDS-VID \cite{wang2014person}, PRID2011 \cite{hirzer2011person} and  P-DESTRE \cite{kumar2020pdestre}), we report the effectiveness of the proposed solution against nine methods considered to represent the state-of-the-art. Our results not only point for consistent improvements in performance over the best baseline, but also the proposed solution can be easily tied to various of the previously published methods, also contributing for improvements in the baseline performance of such techniques.

In summary, we provide the following contributions:
\begin{itemize}
    \item In order to better capture the variations between adjacent frames, we propose a symbolic temporal pooling. Unlike previous temporal pooling methods that use either \emph{max} or \emph{avg} functions to get video level descriptions, the proposed method represents each video in a distribution-valued symbolic form. We extract the frame-level features and represent them in distribution valued symbolic form by fitting a Empirical Cumulative Distribution Function (ECDF).
   \item We introduce a symbolic triplet loss function to address the proposed kind of representation, which enables to obtain the loss between two symbolic objects.
   \item An extensive empirical validation of our solution with respect to the state-of-the-art was conducted, in four publicly available video based re-id datasets: MARS \cite{zheng2016mars}, PRID2011 \cite{hirzer2011person}, iLIDS-VID \cite{wang2014person} and P-DESTRE \cite{kumar2020pdestre}.\\
\end{itemize}

The remainder of this paper is organized as follows: Section~\ref{sec:Related} summarizes the most relevant research in the scope of our work. Section~\ref{sec:ProposedMethod} provides a detailed description of the proposed method. In Section~\ref{sec:Results} we discuss the obtained results and the conclusions are given in Section~\ref{sec:Conclusions}.

\section{Related Work}
\label{sec:Related}
In this section, we briefly review the most relevant research in terms of video-based re-identification methods and symbolic representation techniques. 

\textbf{Video Re-ID:} The current person re-id works can be broadly classified into two groups: image-based \cite{liu2017end,zhao2017deeply,wang2018mancs}  and video-based \cite{gu2019temporal,liu2019,rao2018video}. According to our purposes, in this work we pay special attention to  video based re-id techniques.

\textit{Temporal Aggregation:} Temporal aggregation methods either adopt Temporal Pooling  \cite{zheng2016mars,li2018diversity} or  Recurrent Neural Networks (RNN) \cite{mclaughlin2016recurrent,liu2019spatial,zhang2020ordered} to learn temporal cues across the different frames. Suh \etal \cite{suh2018part} developed a part-aligned bi-linear representation  model,  composed of a two-stream network and a bi-linear pooling layer. Among the two-stream network, one stream extracts the appearance map and another one focuses in the body-part map. Liu \etal \cite{liu2019spatial} proposed a spatial temporal Mutual promotion Model, that starts by using a Refining Recurrent Unit (RRU) to recover the missing parts (ocludded) in a frame. Further, a Spatial-Temporal Clues Integration Module (STIM) is introduced to integrate spatial and temporal information.

\textit{Optical Flow:} In optical flow-based methods, temporal cues are extracted using the classical optical flow technique or its minor variations, which are then fused to spatial information.  McLaughlin \etal \cite{mclaughlin2016recurrent} developed an optical flow and RNN-based model to capture short- and long-term temporal information. Feichtenhofer \etal \cite{feichtenhofer2017spatiotemporal} proposed a two-stream architecture that combines spatial and motion pathways by motion gating.
Chung \etal \cite{chung2017two} proposed a two stream convolutional neural network architecture, where each stream is a Siamese Network used to learn spatial and temporal information separately. Wu \etal \cite{wu2019spatio} proposed a spatio-temporal association representation model that fuses appearance and short-term motion at the local level, and spatial and temporal features at the global level. 

\textit{3D CNN:} These methods use 3D convolutional kernels in the spatial and temporal dimensions\cite{tran2015learning,qiu2017learning}. Gao and Nevatia \cite{gao2018revisiting} proposed a 3D CNN architecture to extract video level features, while Li \etal \cite{li2020multi} using the Multi-Scale 3D (M3D) convolution layer to learn temporal cues. Despite showing promising results, this kind of 3D models are computationally expensive and vulnerable to spatial misalignment.

Various efforts were made to develop temporal attention based solutions. Li \etal  \cite{li2018diversity} developed a spatio-temporal attention model in which the discriminative regions are inferred using a multiple spatial attention scheme. Liu \etal \cite{liu2019} proposed a non-local video attention network (NVAN) that exploits both spatial and temporal features by using non-local attention layers. Gu \etal \cite{gu2019temporal} proposed a temporal knowledge propagation (TKP) method, where the image representations learn frame features to match the video representation network output in a shared feature space. Zhang \etal \cite{zhang2020multi} proposed a Multi-Granularity Reference aided Attentive Feature Aggregation (MG-RAFA) model, that starts by determining the importance of each feature. Next, the affinity of each feature  with respect to all reference elements is obtained and used as weight for aggregation.

\textbf{Symbolic Data Analysis:} In recent years, the concept of symbolic data analysis (SDA) has been receiving growing attention. Symbolic data representations can be seen as an  extension  to  the  traditional  data  types,  that include interval, distribution, periodic or histogram-based features. Applications based in symbolic representations have been extensively reported (e.g., \cite{nagabhushan2002knowledge, guru2007symbolic, de2007fuzzy, guru2010symbolic, harish2010symbolic,alaei2017efficient}). In all these works, the basic premise is that the variations among the data that compose a \emph{sequence} cannot be captured by a \emph{crisp} data type and that symbolic representations can do that in a much more effective way \cite{IEEE-TCASII00:IAVQ, de2015fuzzy, matsui2014comparison,verde2015dimension}.


 \section{Proposed Method}
 \label{sec:ProposedMethod}
 
\subsection{Problem Formulation}

Let ${V_i}$ represent a video sequence composed of $N$ frames, i.e., $V_i=\{I_{i1},I_{i2},...I_{iN}\}$. Given a query video ${V^q}$ and a set of gallery elements $\mathcal{V}^g=\{V^g_1, \ldots, V^g_k\}$ the re-id problem evolves to find an ordered list for the gallery IDs based in a distance function $d(.,.)$ that measures the similarity between the query and the gallery elements. In practice, the query is deemed to correspond to the i$^{th}$ identity of the gallery set \emph{iff}:

\begin{equation}
R_{i}=\arg \; \underset{j}{\min}\;d( \bm{f}^{q}, {{\bm{f}^g_j}}),     
\end{equation}
where $\bm{f}^{q}$ and $\bm{f}^{{V^g_j}}$ represent the query and gallery feature sets, extracted from the corresponding $V^g_k$ elements.

As above stated, our approach comprises three major steps: i) frame-level feature extraction; ii) video representation using symbolic distribution valued functions; and iii) symbolic triplet loss-based learning. 

\subsection{Frame-Level Feature Extraction}  

Frame-level spatial features are extracted using a backbone architecture. Let $V =\{I_1,I_2,\ldots I_N\}$ be a video sequence composed of $N$ frames. By feeding each frame into the shared backbone, we obtain $F=\{\bm{f}_1,\ldots \bm{f}_N\}$ feature vectors, each one composed of $M$ values $\bm{f}_i=\{f_1,\ldots f_M\}$. 

\subsection{Distribution Valued Video Representation} 

This is the key component of our method, and aims at avoiding that frame-level features are aggregated using \textit{max/avg} or similar pooling functions to get video representations, which was previously reported to significantly reduce the amount of discriminating information (e.g., \cite{verde2015dimension,matsui2014comparison,de2015fuzzy}).  Let $\bm{f}_i=\{f_{i1},\ldots, f_{iN} \}$ be the i$^{th}$ feature vector extracted from $N$ frames.  Let $\psi$ denote the  hyper-dimensional distribution of $\bm{f}_i$ values. The corresponding ECDF values can be obtained based in the histogram description $\Acute{F(i)}$, using $H_i$ weighted intervals, where the histogram for the $f_i$ feature is given by:

\begin{equation}
    f_i=[(L_{1i},\pi_{1i}),...,(L_{ri},\pi_{ri}),...,(L_{H_{i}i},\pi_{H_{i}i})],
\end{equation}
where $L$ denotes the bins, $\pi$ represents the frequency associated with each bin and $H_i$ is the number of bins for the i$^{th}$ feature, such that:

$\forall L_{ri} \in T(i)$ where $T(i)=[\min(f_i),\max(f_i)]$.

Using the corresponding histogram description $\Acute{F(i)}$, the empirical cumulative distribution function $\psi_{i}$ for $i^{th}$ feature is given by:
\begin{align}
    \psi_{i}(p)=w_i + (p-\min(f_i)) (\frac{w_{li}-w_{l-1i}}{\max(f_i)-\min(f_i)}). 
\end{align}

After obtaining the symbolic representation of each feature, the corresponding distribution valued video representation are straightforward to obtain $\Psi=\{\psi_1,\ldots \psi_M\}$.

\subsection{Symbolic Triplet Loss}

The triplet loss is one of the most commonly used functions in re-id approaches. In the conventional triplet loss formulation, the distances from the anchor image to a \emph{positive} and a \emph{negative} elements are obtained. During the learning process, the goal is to keep the distances between the anchors and their negative counterparts larger than the corresponding distances between the anchor/positive pairs:

\begin{equation}
    \mathcal{L}_{tri}(i,j,k)=\max(\varrho+d_{ij}-d_{ik},0),
\end{equation}
where $d_{ij}$ denotes the distance between the i$^{th}$ and j$^{th}$ elements, $\varrho$ is a positive margin value and $i$, $j$, $k$ denote the indices of the anchor, positive and negative images. In this conventional formulation, different metrics (e.g., Euclidean or  Cosine distances) can be used to compute the (dis)similarity between feature sets, but these will only work for crisp data \cite{yu2018correcting,cheng2016person,yuan2019defense}. Accordingly, as the proposed method uses a symbolic representation based in the distribution function \emph{per feature}, the conventional triplet loss cannot be applied directly. There are several distance measures that can be used to obtain the distance between two distribution-valued objects \cite{gibbs2002choosing} and, among them, the Wasserstein metric was reported to perform the best,  while also interpreting the characteristics of the distribution \cite{Wasserstein_metric}. Thus, we propose a symbolic triplet loss that uses the Wasserstein metric to obtain the distance between feature vectors.

Let $\psi_i$ and $\psi_j$ represent the multi-dimensional distributions of the i$^{th}$ and j$^{th}$ feature vectors. The corresponding Wasserstein distance is given by:

\begin{equation}
\begin{split}
    D_w(\psi_i, \psi_j) &=\int_{-\infty}^{-\infty}\left | \psi_i(p)-\psi_j(p)  \right | dp \\
                       & = \int_{0}^{1}\left | {\psi_i}^{-1}(t)-{\psi_j}^{-1}(t)  \right | dt,
\end{split}
\label{wast}
\end{equation}
where ${\psi_i}^{-1}(t)$ is the quantile function, given by:
\begin{equation}
{\psi_i}^{-1}(t)=\min(f_i)+ \left(\frac{t-w_{l-1i}}{w_{li}-w_{l-1i}}\right)\Big(\max(f_i)-\min(f_i)\Big).
\end{equation}
Finally, the distance between two distribution-valued video representations (each with $M$ features) is given by:
\begin{equation}
 D_w(\psi_i, \psi_j) = \sum_{m=1}^{M}\sum_{t=1}^{T}    {\psi_{im}}^{-1}(t)-{\psi_{jm}}^{-1}(t).
\end{equation}

\section{Results and Discussion}
\label{sec:Results}

\subsection{Datasets}
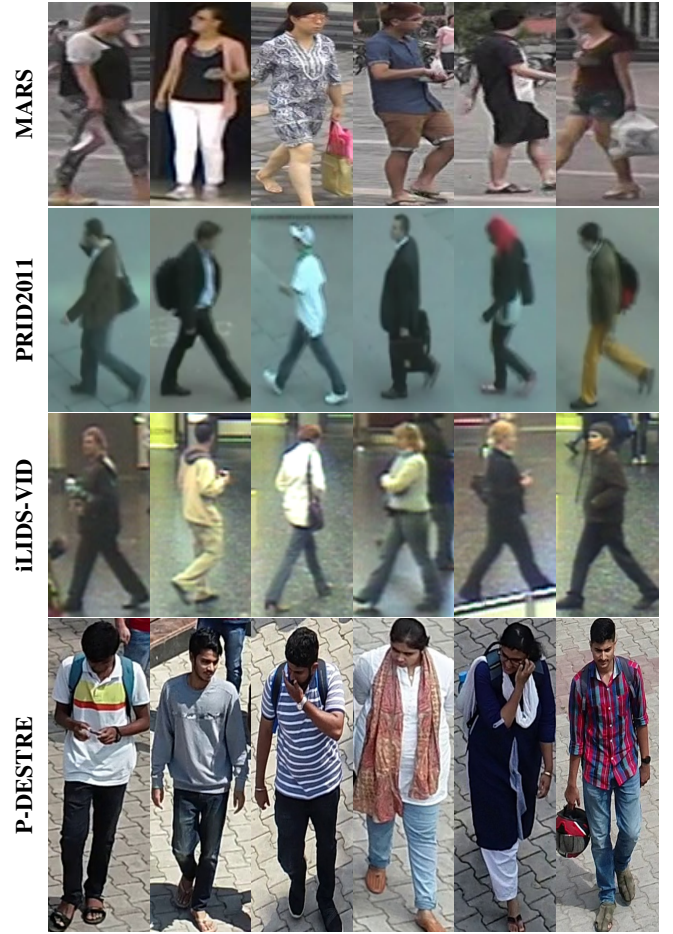
\begin{figure}[ht!]
\begin{center}
\begin{tikzpicture}

\def\sizeImg{1.35}
\def\deltaY{-2.4}

\draw (0*\sizeImg,0) node(segment_ok)  {\includegraphics[width=\sizeImg cm]{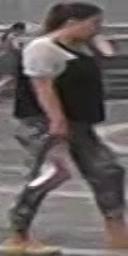}};  	
\draw (1*\sizeImg,0) node(segment_ok)  {\includegraphics[width=\sizeImg cm]{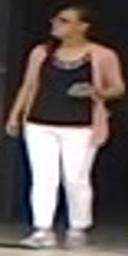}};  	
\draw (2*\sizeImg,0) node(segment_ok)  {\includegraphics[width=\sizeImg cm]{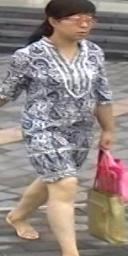}};  	
\draw (3*\sizeImg,0) node(segment_ok)  {\includegraphics[width=\sizeImg cm]{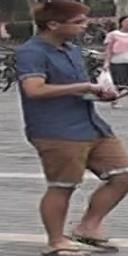}};  	
\draw (4*\sizeImg,0) node(segment_ok)  {\includegraphics[width=\sizeImg cm]{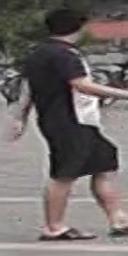}};  	
\draw (5*\sizeImg,0) node(segment_ok)  {\includegraphics[width=\sizeImg cm]{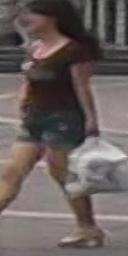}};  	
\draw (-1.0, 0) node[rectangle, rotate=90] {\small{\textbf{MARS}}};  
\draw (0*\sizeImg,0+1.14*\deltaY) node(segment_ok)  {\includegraphics[width=\sizeImg cm]{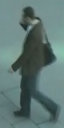}};  	
\draw (1*\sizeImg,0+1.14*\deltaY) node(segment_ok)  {\includegraphics[width=\sizeImg cm]{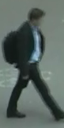}};  	
\draw (2*\sizeImg,0+1.14*\deltaY) node(segment_ok)  {\includegraphics[width=\sizeImg cm]{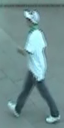}};  	
\draw (3*\sizeImg,0+1.14*\deltaY) node(segment_ok)  {\includegraphics[width=\sizeImg cm]{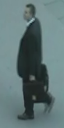}};  	
\draw (4*\sizeImg,0+1.14*\deltaY) node(segment_ok)  {\includegraphics[width=\sizeImg cm]{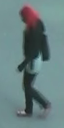}};  	
\draw (5*\sizeImg,0+1.14*\deltaY) node(segment_ok)  {\includegraphics[width=\sizeImg cm]{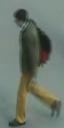}};  	
\draw (-1.0, 0+1.14*\deltaY) node[rectangle, rotate=90] {\small{\textbf{PRID2011}}};  

\draw (0*\sizeImg,0+2.28*\deltaY) node(segment_ok)  {\includegraphics[width=\sizeImg cm]{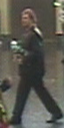}};
\draw (1*\sizeImg,0+2.28*\deltaY) node(segment_ok)  {\includegraphics[width=\sizeImg cm]{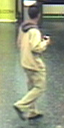}};
\draw (2*\sizeImg,0+2.28*\deltaY) node(segment_ok)  {\includegraphics[width=\sizeImg cm]{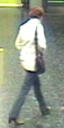}}; 
\draw (3*\sizeImg,0+2.28*\deltaY) node(segment_ok)  {\includegraphics[width=\sizeImg cm]{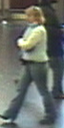}}; 
\draw (4*\sizeImg,0+2.28*\deltaY) node(segment_ok)  {\includegraphics[width=\sizeImg cm]{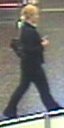}}; 
\draw (5*\sizeImg,0+2.28*\deltaY) node(segment_ok)  {\includegraphics[width=\sizeImg cm]{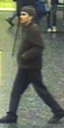}};
\draw (-1.0, 0+2.28*\deltaY) node[rectangle, rotate=90] {\small{\textbf{iLIDS-VID}}}; 

\draw (0*\sizeImg,0+3.72*\deltaY) node(segment_ok)  {\includegraphics[width=\sizeImg cm]{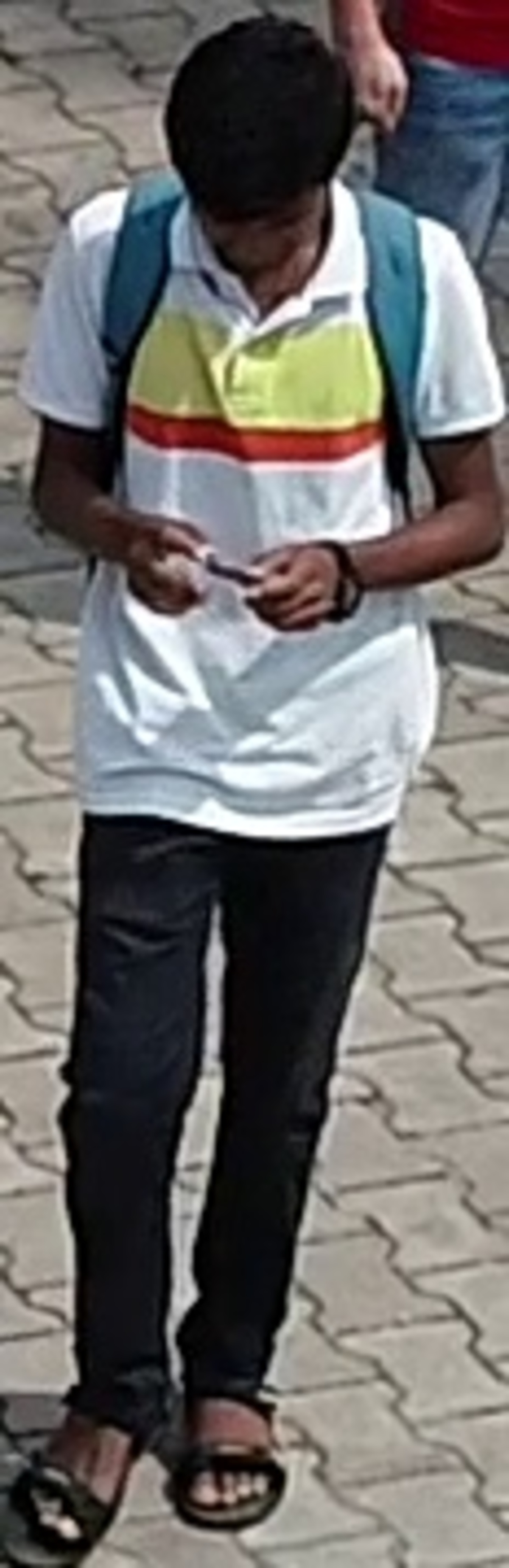}};  	
\draw (1*\sizeImg,0+3.72*\deltaY) node(segment_ok)  {\includegraphics[width=\sizeImg cm]{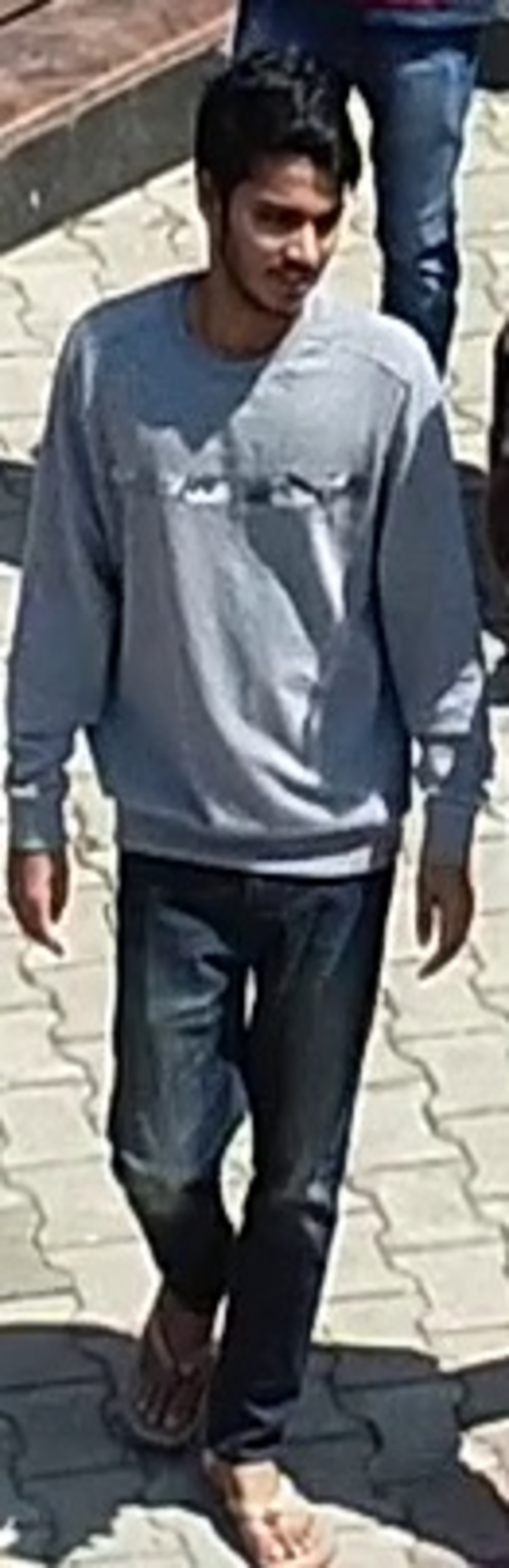}};  	
\draw (2*\sizeImg,0+3.72*\deltaY) node(segment_ok)  {\includegraphics[width=\sizeImg cm]{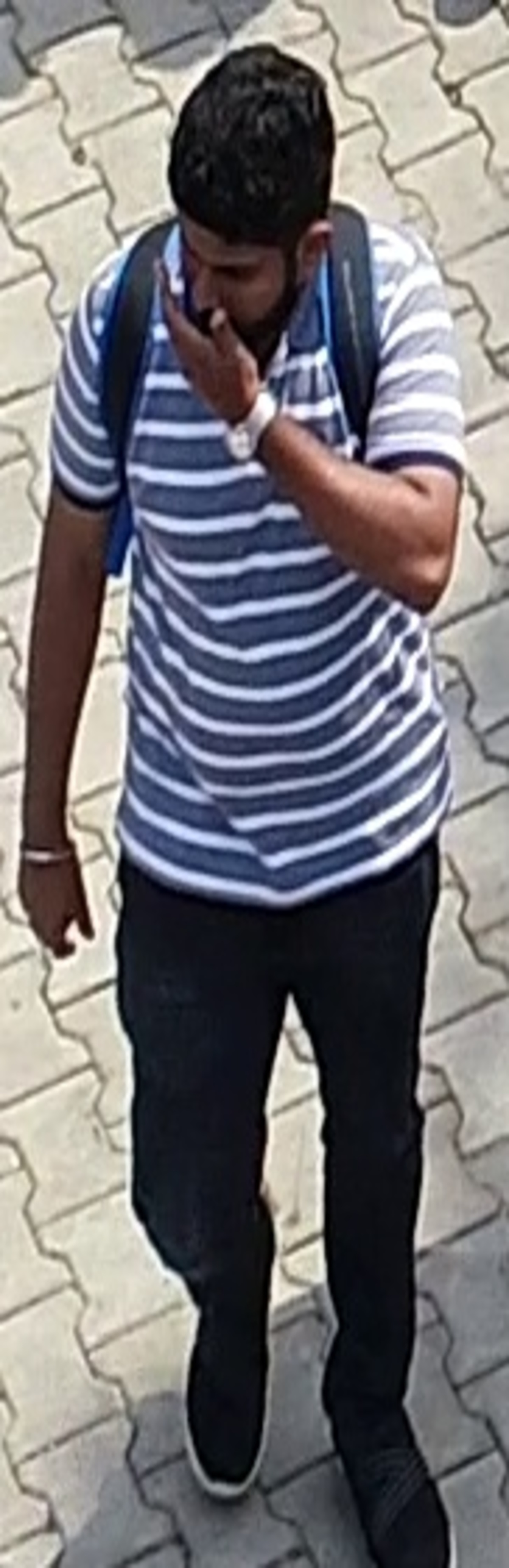}};  	
\draw (3*\sizeImg,0+3.72*\deltaY) node(segment_ok)  {\includegraphics[width=\sizeImg cm]{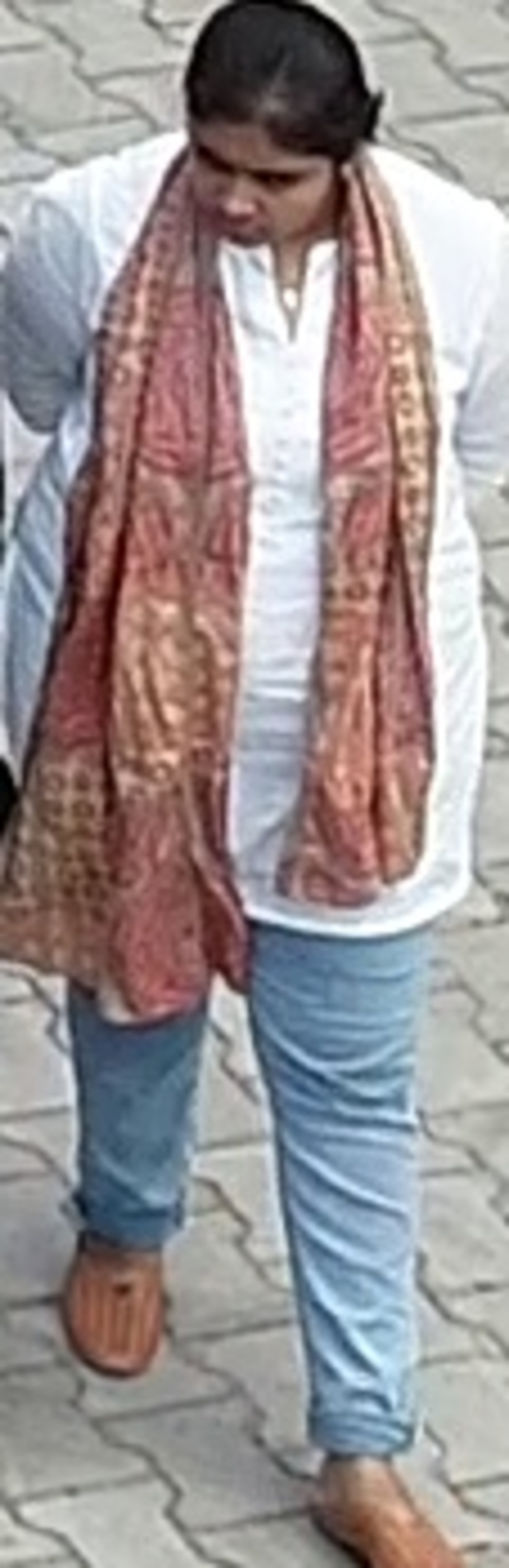}};  	
\draw (4*\sizeImg,0+3.72*\deltaY) node(segment_ok)  {\includegraphics[width=\sizeImg cm]{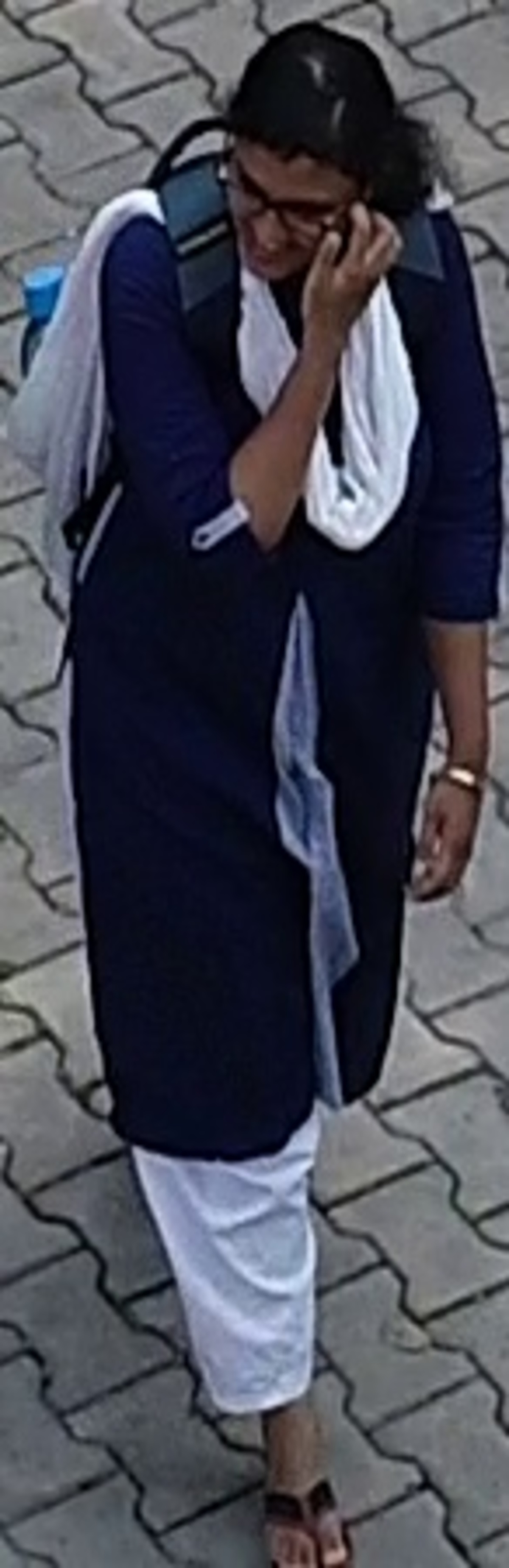}};  	
\draw (5*\sizeImg,0+3.72*\deltaY) node(segment_ok)  {\includegraphics[width=\sizeImg cm]{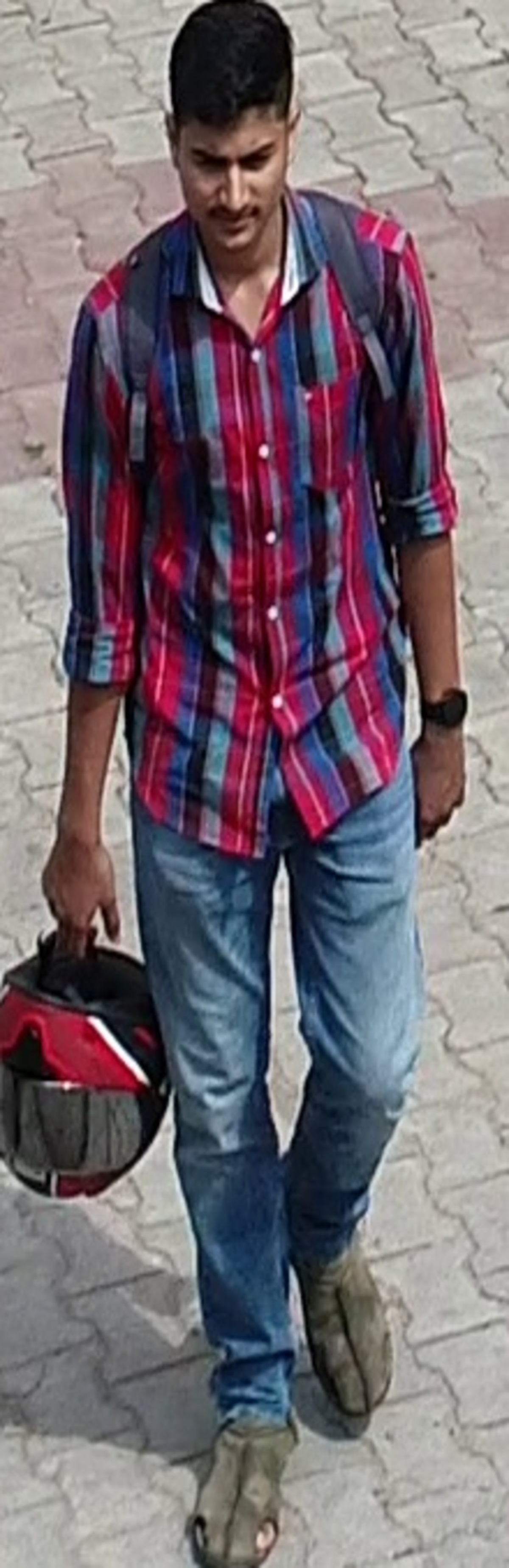}};  	
\draw (-1.0, 0+3.72*\deltaY) node[rectangle, rotate=90] {\small{\textbf{P-DESTRE}}};  
\end{tikzpicture}
    \caption{Sample images from the datasets used in the experimental validation of the proposed method. From top to bottom rows, images of the MARS, PRID2011, iLIDS-VID, and P-DESTRE are shown.}
        \label{fig:Datasets}
    \end{center}
\end{figure}

Our experiments were conducted in four publicly available video-based re-id datasets: MARS \cite{zheng2016mars}, iLIDS-VID \cite{wang2014person}, PRID2011 \cite{hirzer2011person}, and P-DESTRE \cite{kumar2020pdestre}. Examples of the typical images in each set are provided in Fig.~\ref{fig:Datasets}.

\textbf{MARS \cite{zheng2016mars}}. It is the largest video based re-id dataset and was captured inside a university campus, using six surveillance cameras. It comprises 17,503 tracklets of 1,261 IDs, along with  3,248 distractor tracklets. 

\textbf{iLIDS-VID \cite{wang2014person}}. This set was captured in an airport arrival lounge, using two non-overlapping cameras. It comprises image sequences of 300 IDs, where the tracklet length varies between 23 to 192, with an average value of 73.

\textbf{PRID2011  \cite{hirzer2011person}}. It was captured using two non-overlapping cameras, and it comprises a total 749 IDs among which only 200 people appeared in both cameras. The tracklet length varies from 5 to 675 with an average value of 100 images/sequence.

\textbf{P-DESTRE \cite{kumar2020pdestre}}. In opposition to all the other datasets used (captured using stationary cameras), this is an UAV-based dataset, specifically devoted for pedestrian analysis. It comprises 1,894 tracklets of 608 IDs, with an average length of 674 images per sequences. 

\subsection{Experimental Setting and Evaluation Metrics}
\label{ssec:Datasets}
In this section, we report an extensive comparison between the effectiveness of our solution with respect to the state-of-the-art solutions. Considering that re-id is currently a crowded topic, nine techniques were selected as baselines: 

 \begin{itemize}
\item Chung \etal~\cite{chung2017two} : This method\footnote{\scriptsize\color{blue}https://github.com/icarofua/siamese-two-stream} uses a Siamese network as backbone, temporal pooling as aggregation and a \textit{softmax} as the loss function. Interestingly, it is the unique method considered state-of-the-art that learns the spatial and temporal features separately. In our experiments, the best results of this model were obtained when using stochastic gradient descent method with the following parameterization: epochs=1000, learning rate=0.0001 and  momentum=0.9. The Siamese cost function margin was set to 2 and the Euclidean distance was used as similarity metric.

\item Gao and Nevatia\cite{gao2018revisiting}:  This method\footnote{\scriptsize\color{blue} https://github.com/jiyanggao/Video-Person-ReID} uses 3D ResNet which uses 3D convolution kernels as backbone architecture and \textit{triplet} as a loss function. It uses RNN and 3D CNN models to obtain the video level representations. The model was trained using Adam optimizer with parameters: epoch=800, learning rate=0.0003, weight decay ${5\rm e}^{-4}$, Triplet loss margin=0.3. During testing, Euclidean distance is used as a metric to compute the similarity between query and gallery.

\item Rao \etal\cite{rao2018video}:  This method\footnote{\scriptsize\color{blue} https://github.com/rshivansh/Spatial-Temporal-attention} uses a siamese network, \textit{hinge} loss and \textit{softmax} loss functions. Also, it proposes a spatial-temporal attention mechanism, based on the attention scores. The model was trained using stochastic gradient descent with learning rate equal to 0.0001. The margin for hinge loss was set to 2 and 16 frames were considered to represent a sequences in the training phase. During testing, the Euclidean distance was used as a metric to obtain the similarity any pair of elements.

\item Spatio Temporal Attention Model (STAM)~\cite{li2018diversity}: This method\footnote{\scriptsize \color{blue} https://github.com/ShuangLI59/Diversity-Regularized-Spatiotemporal-Attention} uses the \emph{ResNet-50} architecture as backbone, temporal attention as aggregation and an Online Instance Matching Loss function. Essentially, it discriminates between the different parts of the body using spatial-temporal attention. Initially, features are extracted from the local region are organized using the spatial criterion and further combined using temporal attention. This model was trained using stochastic gradient descent method with epoch=800, learning rate=0,1 and it is then dropped to 0.01. During training Restricted Random Sampling is used to select training samples. During testing, the first image from each of N  segments as a testing sample and its L2-normalized features are utilized to compute the similarity between query and  gallery.

\item Temporal Knowledge Propagation (TKP)~\cite{gu2019temporal}:  This model\footnote{\scriptsize\color{blue} https://github.com/guxinqian/TKP } uses \emph{ResNet-50} as backbone model. The input frames were resized to $256\times128$. The model was trained using Adam optimizer with following parameters: learning rate=0.0003, 150 epochs, learning rate decay=0.1, weight decay=${5\rm e}^{-4}$. TKP uses four-loss functions: classification loss, Triplet loss, Feature-based TKP loss, and Distance-based TKP loss. During testing, the Euclidean distance was used as similarity measure.

\item Spatial Temporal Mutual Promotion Model (STMPM)~\cite{liu2019spatial}: This method\footnote{\scriptsize\color{blue} https://github.com/yolomax/rru-reid } uses the \emph{Inception-v3} architecture as backbone model and cross-entropy as a loss function. It handles  occlusions by recovering the missing parts and uses a 3D-CNN and global average pooling for temporal aggregation. This model was trained using  stochastic gradient descent while number of epochs, learning rate, and momentum set to 800, 0.01, 0.9 respectively. Model parameters N, K and T and m were set to 10, 2, 8 and 0.4, respectively. During test, the cosine distance is used as a metric.

\item CoSAM~\cite{subramaniam2019co}:
This model\footnote{\scriptsize\color{blue} https://github.com/InnovArul/vidreid\_cosegmentation} uses the SE-ResNet50 architecture as backbone and extracts salient spatial features by a co-segmentation based attention model. The input frames were resized to $256 \times 128$, and the model was trained using the Adam optimizer, learning rate 0.0001 and weight decay ${5\rm e}^{-4}$. During test, the Euclidean distance is used as similarity metric.

\item GLTR~\cite{li2019global}: This model\footnote{\scriptsize\color{blue} https://github.com/kanei1024/GLTR} uses ResNet50 as backbone architecture. The input frames were resized to $256\times128$, and the model was trained using the stochastic gradient descent algorithm 400 epochs, learning rate 0.01, momentum 0.9 and weight decay ${5\rm e}^{-4}$. For DTP and TSA training, 16  adjacent frames compose each sequence and used as input.  During test, the Euclidean distance is used as similarity metric.

\item  Non-local Video Attention Network (NVAN)~\cite{liu2019}: This method\footnote{\scriptsize\color{blue} https://github.com/jackie840129/STE-NVAN } uses ResNet50 as a backbone network and plugs two non-local attention layer after Conv3\_3, and Conv3\_4 convolutional layers and three non-local layers at Conv4\_4, Conv4\_5, and Conv4\_6. It incorporates both spatial and temporal features using the non-local attention operation at multiple feature levels and uses triplet and cross entropy loss functions. The input frames are resized into $256\times 128$. The model was trained with the Adam algorithm,  with 300 epochs, learning rate=0.0001. During test, the Euclidean distance is used as similarity metric.
 \end{itemize}

For the PRID2011 and iLIDS-VID sets we strictly followed the evaluation protocols described in \cite{li2018diversity, hirzer2011person}. For the other sets (MARS and P-DESTRE), we used the evaluation protocols described in \cite{zheng2016mars} and \cite{kumar2020pdestre}, respectively.
As performance measures, two widely used evaluation metrics were chosen: the Cumulative Matching Characteristics (CMC) and the Mean Average Precision (mAP) values. Due to the single gallery instance per probe feature of the iLIDS-VID and PRID2011 sets, only the CMC values were used in these cases.

\subsection{Backbone Architectures}

Two well-known re-id backbone architectures were used to extract frame-level spatial features: the \emph{ResNet50} and \emph{SE-ResNet50}. We report the performance of these models when using the proposed temporal pooling method, in comparison to the results obtained when using the average pooling method. 

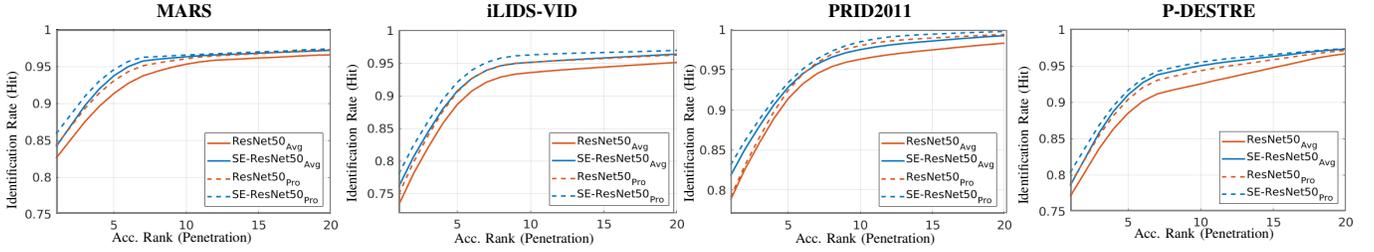
\begin{figure*}
    \centering
    \resizebox{1\textwidth}{!}{%
\tikzset{every picture/.style={line width=0.75pt}} 

\begin{tikzpicture}[x=0.75pt,y=0.75pt,yscale=-1,xscale=1]

\draw (160.5,132.5) node  {\includegraphics[width=198.75pt,height=129.75pt]{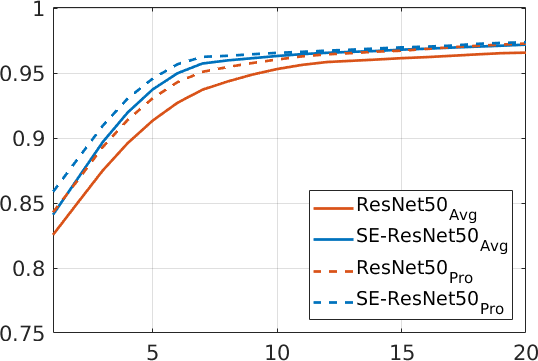}};
\draw (453,132.5) node  {\includegraphics[width=201pt,height=128.25pt]{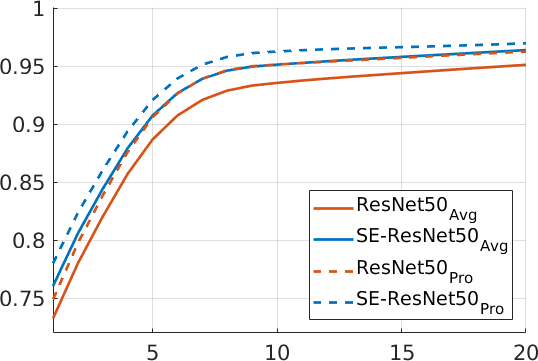}};
\draw  [color={rgb, 255:red, 128; green, 128; blue, 128 }  ,draw opacity=1 ][line width=0.75]  (55,50) -- (287,50) -- (287,205) -- (55,205) -- cycle ;
\draw  [color={rgb, 255:red, 128; green, 128; blue, 128 }  ,draw opacity=1 ][line width=0.75]  (346,50) -- (579,50) -- (579,205) -- (346,205) -- cycle ;
\draw (734,132.5) node  {\includegraphics[width=198pt,height=129.75pt]{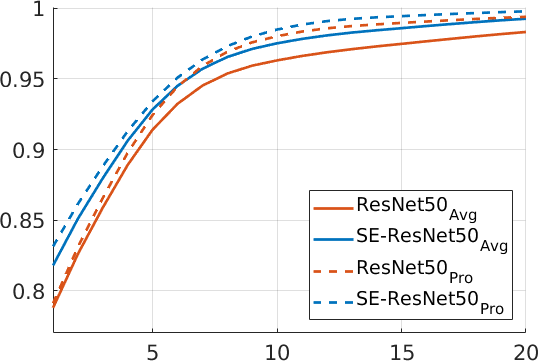}};
\draw  [color={rgb, 255:red, 128; green, 128; blue, 128 }  ,draw opacity=1 ][line width=0.75]  (628,50) -- (862,50) -- (862,205) -- (628,205) -- cycle ;
\draw (1023,131.5) node  {\includegraphics[width=199.5pt,height=128.25pt]{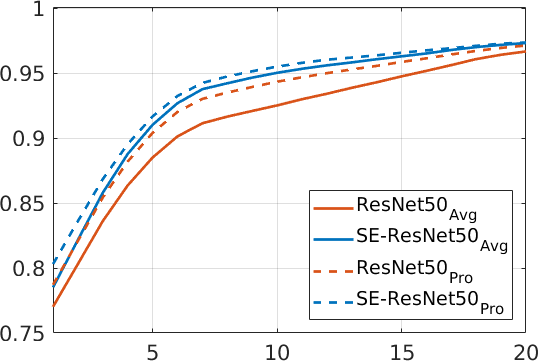}};
\draw  [color={rgb, 255:red, 128; green, 128; blue, 128 }  ,draw opacity=1 ][line width=0.75]  (917,49) -- (1150,49) -- (1150,203) -- (917,203) -- cycle ;

\draw (10.5,203.5) node [anchor=north west][inner sep=0.75pt]  [color={rgb, 255:red, 0; green, 0; blue, 0 }  ,opacity=1 ,rotate=-270] [align=left] {\textcolor[rgb]{0,0,0}{{\small Identification Rate (Hit)}}};
\draw (100,220) node [anchor=north west][inner sep=0.75pt]   [align=left] {\textcolor[rgb]{0,0,0}{{\small Acc. Rank (Penetration)}}};
\draw (138,26) node [anchor=north west][inner sep=0.75pt]   [align=left] {\textbf{\textcolor[rgb]{0,0,0}{{\large MARS}}}};
\draw (299.5,197.5) node [anchor=north west][inner sep=0.75pt]  [color={rgb, 255:red, 0; green, 0; blue, 0 }  ,opacity=1 ,rotate=-270] [align=left] {\textcolor[rgb]{0,0,0}{{\small Identification Rate (Hit)}}};
\draw (419,26) node [anchor=north west][inner sep=0.75pt]   [align=left] {\textcolor[rgb]{0,0,0}{\textbf{{\large iLIDS-VID}}}};
\draw (709,26) node [anchor=north west][inner sep=0.75pt]   [align=left] {\textbf{\textcolor[rgb]{0,0,0}{{\large PRID2011}}}};
\draw (993,26) node [anchor=north west][inner sep=0.75pt]   [align=left] {\textbf{\textcolor[rgb]{0,0,0}{{\large P-DESTRE}}}};
\draw (391,221) node [anchor=north west][inner sep=0.75pt]   [align=left] {\textcolor[rgb]{0,0,0}{{\small Acc. Rank (Penetration)}}};
\draw (681,221) node [anchor=north west][inner sep=0.75pt]   [align=left] {\textcolor[rgb]{0,0,0}{{\small Acc. Rank (Penetration)}}};
\draw (969,217) node [anchor=north west][inner sep=0.75pt]   [align=left] {\textcolor[rgb]{0,0,0}{{\small Acc. Rank (Penetration)}}};
\draw (586.5,197.5) node [anchor=north west][inner sep=0.75pt]  [color={rgb, 255:red, 0; green, 0; blue, 0 }  ,opacity=1 ,rotate=-270] [align=left] {\textcolor[rgb]{0,0,0}{{\small Identification Rate (Hit)}}};
\draw (870.5,199.5) node [anchor=north west][inner sep=0.75pt]  [color={rgb, 255:red, 0; green, 0; blue, 0 }  ,opacity=1 ,rotate=-270] [align=left] {\textcolor[rgb]{0,0,0}{{\small Identification Rate (Hit)}}};
\end{tikzpicture}
}
 \caption{Comparison between the CMC curves
observed in the four datasets, when using the temporal "avg" function and the proposed method ("pro"). The solid orange and cyan lines correspond the ResNet50+Avg and SE-ResNet50+Avg results. The dotted orange and cyan lines provide the corresponding ResNet50+Pro and SE-ResNet50+Pro results.}
\label{res_seres_g}
\end{figure*}

The performance values are summarized in Table \ref{backbone}. Also, Fig. \ref{res_seres_g} provides the top 20 CMC values for the four datasets, where the horizontal axes denote the rank and the vertical axes provide the corresponding identification rate. The solid lines (orange and cyan) regard the values obtained when using the classical $avg$ pooling function (underlined "avg" series), whereas the dashed lines correspond to the same architectures coupled to the method proposed in this paper (underlined "pro" series).

From the analysis of  Table \ref{backbone} and Fig. \ref{res_seres_g}, it is evident that the proposed temporal pooling method consistently outperforms the $avg$ pooling method, on both models and all datasets. Among the two backbone architectures, the SE-ResNet50 model got consistently better results than its ResNet50 counterpart. In particular, the proposed temporal pooling method improved the mAP values by 0.77 \% (ResNet50) and 1.56\% (SE-ResNet50) in the MARS set. For the P-DESTRE dataset, the mAP improvements were about 0.77\% and 1.79\% for the ResNet50 and SE-ResNet50. Regarding the CMC rank-1 values, we observed improvements of 1.74\% (ResNet50) \& 1.74\% (SE-ResNet50) in MARS, 1.63\% (ResNet50) \& 1.94 (SE-ResNet50) in iLIDS-VID, 0.33\% (ResNet50) \& 1.34\% (SE-ResNet50) in PRID2011, 1.70\% (ResNet50) \& 1.80 \% (SE-ResNet50) in P-DESTRE. Similar improvements were observed for CMC rank-20 values, of  0.7\% (ResNet50) \& 0.2  \% (SE-ResNet50) were observed in MARS, 1.10 \% (ResNet50) \& 0.55 (SE-ResNet50) in iLIDS-VID, 1.10 \% and (ResNet50) \& 0.50\% (SE-ResNet50) in PRID2011,  0.44\% (ResNet50) \& 0.20\% (SE-ResNet50) in the P-DESTRE set.

\begin{figure*}[!]
    \centering
    \resizebox{1\textwidth}{!}{%
\tikzset{every picture/.style={line width=0.75pt}} 

\begin{tikzpicture}[x=0.75pt,y=0.75pt,yscale=-1,xscale=1]

\draw (160.5,132.5) node  {\includegraphics[width=198.75pt,height=129.75pt]{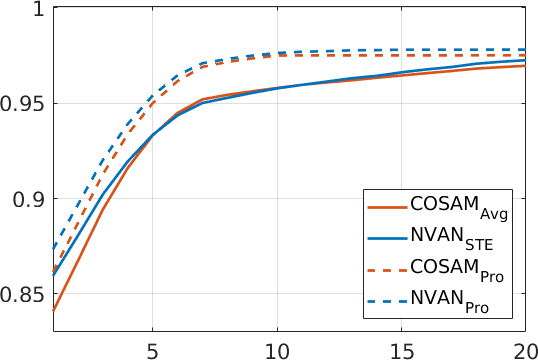}};
\draw (453,132.5) node  {\includegraphics[width=201pt,height=128.25pt]{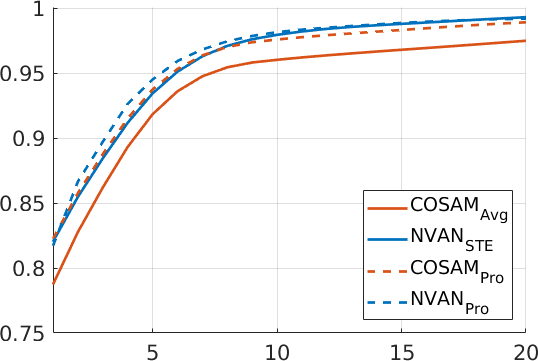}};
\draw  [color={rgb, 255:red, 128; green, 128; blue, 128 }  ,draw opacity=1 ][line width=0.75]  (55,50) -- (287,50) -- (287,205) -- (55,205) -- cycle ;
\draw  [color={rgb, 255:red, 128; green, 128; blue, 128 }  ,draw opacity=1 ][line width=0.75]  (346,50) -- (579,50) -- (579,205) -- (346,205) -- cycle ;
\draw (734,132.5) node  {\includegraphics[width=198pt,height=129.75pt]{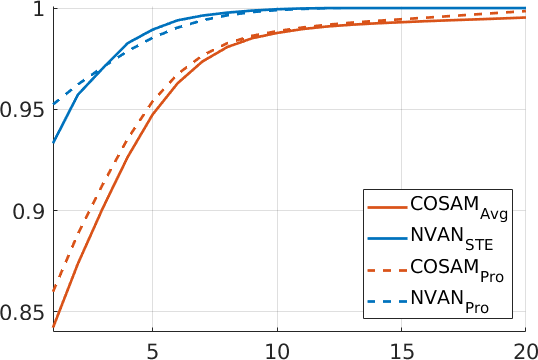}};
\draw  [color={rgb, 255:red, 128; green, 128; blue, 128 }  ,draw opacity=1 ][line width=0.75]  (628,50) -- (862,50) -- (862,205) -- (628,205) -- cycle ;
\draw (1023,131.5) node  {\includegraphics[width=199.5pt,height=128.25pt]{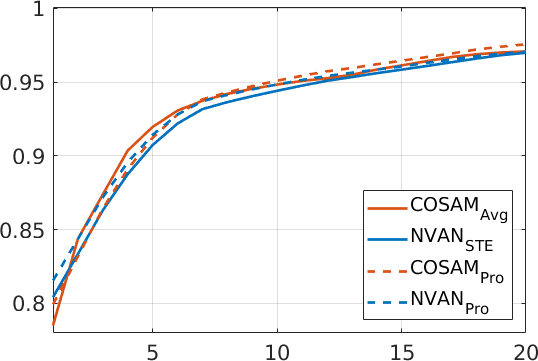}};
\draw  [color={rgb, 255:red, 128; green, 128; blue, 128 }  ,draw opacity=1 ][line width=0.75]  (917,49) -- (1150,49) -- (1150,203) -- (917,203) -- cycle ;

\draw (10.5,203.5) node [anchor=north west][inner sep=0.75pt]  [color={rgb, 255:red, 0; green, 0; blue, 0 }  ,opacity=1 ,rotate=-270] [align=left] {\textcolor[rgb]{0,0,0}{{\small Identification Rate (Hit)}}};
\draw (100,220) node [anchor=north west][inner sep=0.75pt]   [align=left] {\textcolor[rgb]{0,0,0}{{\small Acc. Rank (Penetration)}}};
\draw (138,26) node [anchor=north west][inner sep=0.75pt]   [align=left] {\textbf{\textcolor[rgb]{0,0,0}{{\large MARS}}}};
\draw (299.5,197.5) node [anchor=north west][inner sep=0.75pt]  [color={rgb, 255:red, 0; green, 0; blue, 0 }  ,opacity=1 ,rotate=-270] [align=left] {\textcolor[rgb]{0,0,0}{{\small Identification Rate (Hit)}}};
\draw (419,26) node [anchor=north west][inner sep=0.75pt]   [align=left] {\textcolor[rgb]{0,0,0}{\textbf{{\large iLIDS-VID}}}};
\draw (709,26) node [anchor=north west][inner sep=0.75pt]   [align=left] {\textbf{\textcolor[rgb]{0,0,0}{{\large PRID2011}}}};
\draw (993,26) node [anchor=north west][inner sep=0.75pt]   [align=left] {\textbf{\textcolor[rgb]{0,0,0}{{\large P-DESTRE}}}};
\draw (391,221) node [anchor=north west][inner sep=0.75pt]   [align=left] {\textcolor[rgb]{0,0,0}{{\small Acc. Rank (Penetration)}}};
\draw (681,221) node [anchor=north west][inner sep=0.75pt]   [align=left] {\textcolor[rgb]{0,0,0}{{\small Acc. Rank (Penetration)}}};
\draw (969,217) node [anchor=north west][inner sep=0.75pt]   [align=left] {\textcolor[rgb]{0,0,0}{{\small Acc. Rank (Penetration)}}};
\draw (586.5,197.5) node [anchor=north west][inner sep=0.75pt]  [color={rgb, 255:red, 0; green, 0; blue, 0 }  ,opacity=1 ,rotate=-270] [align=left] {\textcolor[rgb]{0,0,0}{{\small Identification Rate (Hit)}}};
\draw (870.5,199.5) node [anchor=north west][inner sep=0.75pt]  [color={rgb, 255:red, 0; green, 0; blue, 0 }  ,opacity=1 ,rotate=-270] [align=left] {\textcolor[rgb]{0,0,0}{{\small Identification Rate (Hit)}}};
\end{tikzpicture}
}
  \caption{Comparison between the CMC curves obtained, when using the temporal "avg" function and the symbolic representation proposed in this paper ("pro"). The solid orange and cyan lines correspond to the COSAM+Avg and NVAN+STE results. The dotted orange and cyan  lines provide the corresponding COSAM+Pro and NVAN+Pro results.}
\label{cosam_nvan_g}
\end{figure*}

\begin{table*}[!]
\caption{Summary performance of the backbone feature extractors when using the temporal "avg" and the symbolic representation proposed in this paper ("pro").}
\label{backbone}
\resizebox{\textwidth}{!}{%
\begin{tabular}{c|c|c|c|c|c|c|c|c|c|c}
\hline
\multirow{2}{*}{\scriptsize{Backbone}} & \multicolumn{3}{c|}{\scriptsize{MARS}} & \multicolumn{2}{c|}{\scriptsize{iLIDS-VID}} & \multicolumn{2}{c|}{\scriptsize{PRID2011}} & \multicolumn{3}{c}{\scriptsize{P-DESTRE}} \\ \cline{2-11} 
                          & \scriptsize{mAP}     & \scriptsize{R1}     & \scriptsize{R20}    & \scriptsize{R1}            & \scriptsize{R20   }         & \scriptsize{R1 }           & \scriptsize{R20 }          & \scriptsize{mAP}      & \scriptsize{R1}      & \scriptsize{R20}     \\ \hline
\scriptsize{ResNet50+Avg} &
 \scriptsize{75.54} \tiny{$\pm$   3.66} &
\scriptsize{82.50} \tiny{$\pm$1.96} &
 \scriptsize{96.60} \tiny{$\pm$   1.18} &
 \scriptsize{73.20}  \tiny{$\pm$   5.24} &
\scriptsize{95.10}  \tiny{$\pm$   1.85} &
 \scriptsize{78.72} \tiny{$\pm$   4.27} &
 \scriptsize{98.30}  \tiny{$\pm$   2.37} &
 \scriptsize{77.61}  \tiny{$\pm$   10.69} &
  \scriptsize{76.96}  \tiny{$\pm$   13.85} &
 \scriptsize{96.90}  \tiny{$\pm$   2.18} \\ \hline
\scriptsize{SE-ResNet50+Avg} &
 \scriptsize{78.12}  \tiny{$\pm$   3.41} &
 \scriptsize{84.06} \tiny{$\pm$   1.87} &
 \scriptsize{97.20}  \tiny{$\pm$   0.98} &
\scriptsize{76.01 } \tiny{$\pm$   4.62} &
 \scriptsize{96.35}  \tiny{$\pm$   1.50} &
\scriptsize{81.74}  \tiny{$\pm$   3.68} &
 \scriptsize{99.30}  \tiny{$\pm$   2.24} &
 \scriptsize{79.96}  \tiny{$\pm$   10.51} &
 \scriptsize{78.46}  \tiny{$\pm$   13.82} &
 \scriptsize{97.58}  \tiny{$\pm$   2.02} \\ \hline
\scriptsize{ResNet50+Pro} &
\scriptsize{76.31}  \tiny{$\pm$   2.89}&
\scriptsize{84.24} \tiny{$\pm$   1.67}&
\scriptsize{97.30}  \tiny{$\pm$   1.01} &
\scriptsize{74.83}  \tiny{$\pm$   4.32}&
 \scriptsize{96.20}  \tiny{$\pm$   1.64}&
 \scriptsize{79.05 } \tiny{$\pm$   3.79}&
\scriptsize{99.40  }\tiny{$\pm$   2.21} &
 \scriptsize{78.38  }\tiny{$\pm$   10.68} &
 \scriptsize{78.66 } \tiny{$\pm$   13.17}&
\scriptsize{97.34}  \tiny{$\pm$   2.07} \\ \hline
\scriptsize{SE-ResNet50+Pro} &
\scriptsize{79.68} \tiny{$\pm$   2.85} &
\scriptsize{85.80}  \tiny{$\pm$   1.62} &
 \scriptsize{97.40 }\tiny{$\pm$   0.87}&
\scriptsize{77.95 } \tiny{$\pm$   4.19} &
\scriptsize{96.90}  \tiny{$\pm$   1.43}&
\scriptsize{83.08}  \tiny{$\pm$   3.54}&
\scriptsize{99.80}  \tiny{$\pm$   2.25}&
\scriptsize{81.75} \tiny{$\pm$   10.53} &
\scriptsize{80.26}  \tiny{$\pm$   13.34}&
\scriptsize{97.66}  \tiny{$\pm$   1.99}\\ \hline
\end{tabular}%
}
\end{table*}

At a second phase, we assessed the improvements in performance yielding from using the kind of symbolic representations described in this paper in two well known re-id models: the COSAM \cite{subramaniam2019co} and NVAN \cite{liu2019}. Table \ref{cosam_nvan} summarizes the obtained values and Fig. \ref{cosam_nvan_g} provides the top-20 CMC values for the four datasets. Again, the horizontal axes represent the rank and the vertical  axes provide the identification rates. The solid orange and cyan lines regard the original COSAM/NVAN techniques (COSAM+Avg and NVAN+STE) while the dotted lines provide the corresponding values when our proposed symbolic representation were coupled to these methods (COSAM+Pro and NVAN+Pro).

Upon the analysis of Table \ref{cosam_nvan} and Fig. \ref{cosam_nvan_g}, we observe that the proposed temporal pooling method contributed again for improvements in performance. Among both techniques, NVAN consistently outperformed COSAM. In particular, the proposed temporal pooling method improves the mAP by 0.77\% (COSAM) and 0.93\% (NVAN) in MARS Dataset. In the P-DESTRE dataset, the mAP improvement of 1.07\% and 1.43 \% is observed for COSAM and NVAN. Regarding the CMC rank-1, we observe the improvement of 2.11\% (COSAM) \& 1.36\% (NVAN) in MARS, 3.51\% (COSAM) \& 0.32 (NVAN) in iLIDS-VID, 1.78\% (COSAM) \& 1.92\% (NVAN) in PRID2011, 1.42\% (COSAM) \& 2.32 \% (NVAN) in P-DESTRE. Similarly, we observe an improvement of  0.54\%  (for both COSAM and NVAN) in MARS, 1.43\% (COSAM) \& 0.11\% (NVAN) in iLIDS-VID, 0.35\% (COSAM) in PRID2011,  1.44\% (COSAM) \& 1.20\% (NVAN) in the P-DESTRE for CMC rank-20.

\begin{table*}[!]
\caption{Comparison between the  effectiveness of the state-of-the-art video based re-id baseline models in the MARS, iLIDS-VID, PROD2011 and P-DESTRE datasets.}
\label{cosam_nvan}
\resizebox{\textwidth}{!}{%
\begin{tabular}{c|c|c|c|c|c|c|c|c|c|c}
\hline
\multirow{2}{*}{\scriptsize{Backbone}} &
  \multicolumn{3}{c|}{\scriptsize{MARS}} &
  \multicolumn{2}{c|}{\scriptsize{iLIDS-VID}} &
  \multicolumn{2}{c|}{\scriptsize{PRID2011}} &
  \multicolumn{3}{c}{\scriptsize{P-DESTRE}} \\ \cline{2-11} 
 &
\scriptsize{mAP }&
\scriptsize{R1} &
\scriptsize{R20} &
\scriptsize{R1} &
\scriptsize{R20} &
\scriptsize{R1 }&
\scriptsize{R20 }&
\scriptsize{mAP} &
\scriptsize{R1} &
\scriptsize{R20} \\ \hline
\scriptsize{COSAM+Avg} \cite{subramaniam2019co} &
 \scriptsize{78.35}\tiny{$\pm$} 1.66 &
 \scriptsize{84.03}\tiny{$\pm$} 0.91 &
 \scriptsize{96.98} \tiny{$\pm$} 0.98 &
 \scriptsize{78.74}\tiny{$\pm$} 4.10 &
 \scriptsize{97.50}\tiny{$\pm$} 1.31 &
 \scriptsize{84.21}\tiny{$\pm$} 3.07 &
\scriptsize{99.51} \tiny{$\pm$} 0.54 &
\scriptsize{80.42} \tiny{$\pm$} 9.91 &
 \scriptsize{79.14} \tiny{$\pm$} 12.43 &
 \scriptsize{97.10}\tiny{$\pm$} 1.85 \\ \hline
\scriptsize{NVAN+STE} \cite{liu2019} &
\scriptsize{81.13} \tiny{$\pm$} 1.57 &
\scriptsize{85.94} \tiny{$\pm$} 0.94 &
\scriptsize{97.26} \tiny{$\pm$} 0.97 &
\scriptsize{81.71}\tiny{$\pm$} 3.27 &
\scriptsize{99.21}\tiny{$\pm$} 0.99 &
 \scriptsize{93.32} \tiny{$\pm$} 2.54 &
\scriptsize{100.00} \tiny{$\pm$} 0 &
\scriptsize{82.78} \tiny{$\pm$} 10.35 &
\scriptsize{80.36} \tiny{$\pm$} 12.38 &
\scriptsize{97.10} \tiny{$\pm$} 1.93 \\ \hline
\scriptsize{COSAM+Pro} &
\scriptsize{79.12} \tiny{$\pm$} 1.35 &
\scriptsize{86.14} \tiny{$\pm$} 0.84 &
\scriptsize{97.52}\tiny{$\pm$} 0.91 &
\scriptsize{82.25}\tiny{$\pm$} 3.16 &
\scriptsize{98.93}\tiny{$\pm$} 1.27 &
\scriptsize{85.99} \tiny{$\pm$} 2.98 &
\scriptsize{99.86} \tiny{$\pm$} 0.48 &
\scriptsize{81.49} \tiny{$\pm$} 9.64 &
\scriptsize{80.56} \tiny{$\pm$} 11.72 &
\scriptsize{98.54} \tiny{$\pm$} 1.78 \\ \hline
\scriptsize{NVAN+Pro} &
\scriptsize{82.06} \tiny{$\pm$} 1.53 &
\scriptsize{87.30} \tiny{$\pm$} 0.89 &
\scriptsize{97.80} \tiny{$\pm$} 0.84 &
\scriptsize{82.03}\tiny{$\pm$} 3.54 &
\scriptsize{99.32}\tiny{$\pm$} 1.14 &
\scriptsize{95.24} \tiny{$\pm$} 1.24 &
 \scriptsize{100} \tiny{$\pm$} 0 &
\scriptsize{84.21} \tiny{$\pm$} 10.37 &
\scriptsize{82.68} \tiny{$\pm$} 11.51 &
\scriptsize{98.30} \tiny{$\pm$} 1.87 \\ \hline
\end{tabular}%
}
\end{table*}

\subsection{State-of-the-art Comparison}

\begin{figure*}[!]
\centering
    \resizebox{1\textwidth}{!}{%
\tikzset{every picture/.style={line width=0.75pt}} 

\begin{tikzpicture}[x=0.75pt,y=0.75pt,yscale=-1,xscale=1]

\draw [color={rgb, 255:red, 189; green, 16; blue, 224 }  ,draw opacity=1 ][line width=2.25]    (166,619) -- (190.44,619.58) -- (208.5,620) ;
\draw [color={rgb, 255:red, 208; green, 2; blue, 27 }  ,draw opacity=1 ][line width=2.25]    (166,640) -- (190.44,640.58) -- (208.5,641) ;
\draw [color={rgb, 255:red, 74; green, 144; blue, 226 }  ,draw opacity=1 ][line width=2.25]    (166,661) -- (190.44,661.58) -- (208.5,662) ;
\draw [color={rgb, 255:red, 190; green, 88; blue, 9 }  ,draw opacity=1 ][line width=2.25]    (385,621) -- (409.44,621.58) -- (427.5,622) ;
\draw [color={rgb, 255:red, 245; green, 166; blue, 35 }  ,draw opacity=1 ][line width=2.25]    (385,641) -- (409.44,641.58) -- (427.5,642) ;
\draw [color={rgb, 255:red, 144; green, 19; blue, 254 }  ,draw opacity=1 ][line width=2.25]    (385,662) -- (409.44,662.58) -- (427.5,663) ;
\draw [color={rgb, 255:red, 0; green, 0; blue, 0 }  ,draw opacity=1 ][line width=2.25]    (576,620) -- (593.56,620.41) -- (600.44,620.58) -- (618.5,621) ;
\draw [color={rgb, 255:red, 152; green, 235; blue, 62 }  ,draw opacity=1 ][line width=2.25]    (576,639) -- (593.56,639.41) -- (600.44,639.58) -- (618.5,640) ;
\draw [color={rgb, 255:red, 165; green, 30; blue, 46 }  ,draw opacity=1 ][line width=2.25]    (576,660) -- (593.56,660.41) -- (600.44,660.58) -- (618.5,661) ;
\draw [color={rgb, 255:red, 152; green, 235; blue, 62 }  ,draw opacity=1 ][line width=2.25]  [dash pattern={on 6.75pt off 4.5pt}]  (765,621) -- (782.56,621.41) -- (789.44,621.58) -- (807.5,622) ;
\draw [color={rgb, 255:red, 165; green, 30; blue, 46 }  ,draw opacity=1 ][line width=2.25]  [dash pattern={on 6.75pt off 4.5pt}]  (768,652) -- (785.56,652.41) -- (792.44,652.58) -- (810.5,653) ;
\draw (262,153.5) node  {\includegraphics[width=348.75pt,height=194.25pt]{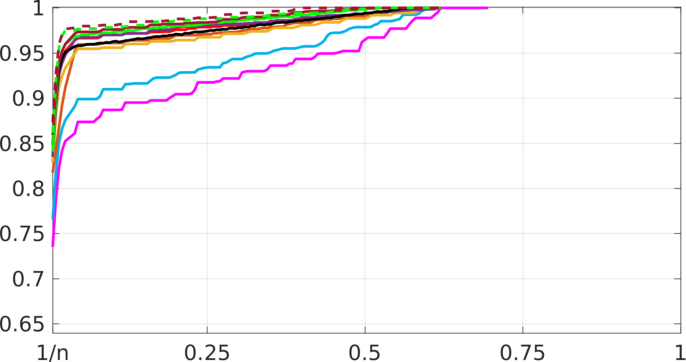}};
\draw (299.5,170.5) node  {\includegraphics[width=284.25pt,height=125.25pt]{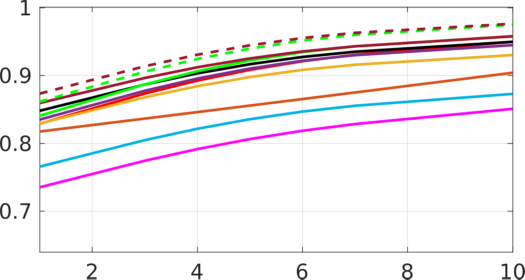}};
\draw (768,159) node  {\includegraphics[width=348.75pt,height=195pt]{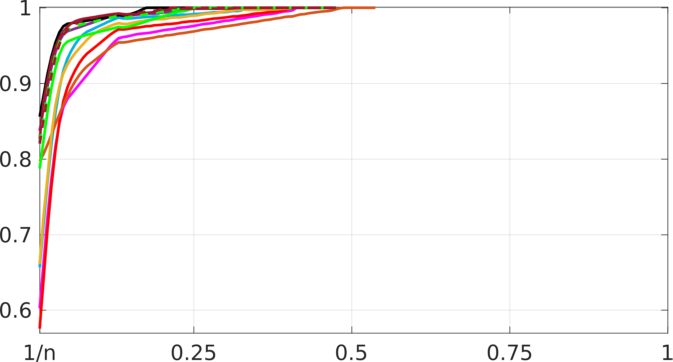}};
\draw (805.5,162) node  {\includegraphics[width=272.25pt,height=136.5pt]{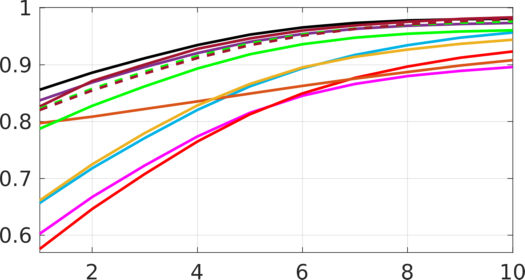}};
\draw (259,451) node  {\includegraphics[width=348.75pt,height=196.5pt]{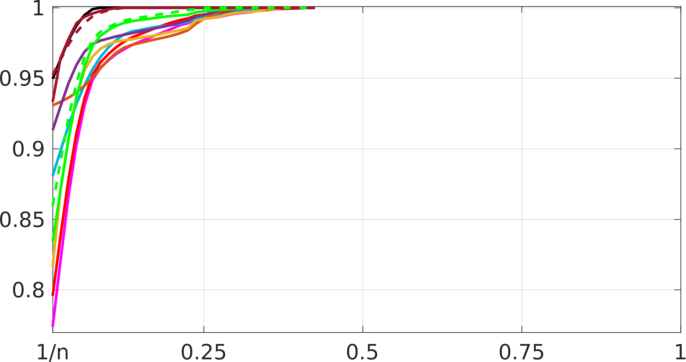}};
\draw (287,463.5) node  {\includegraphics[width=291pt,height=138.75pt]{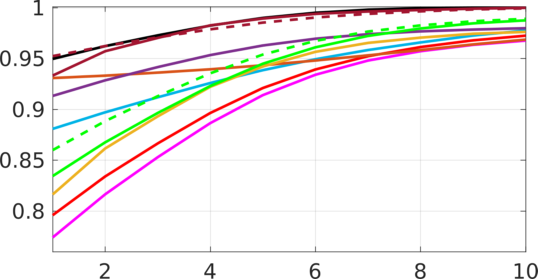}};
\draw (770.5,453.5) node  {\includegraphics[width=349.5pt,height=195.75pt]{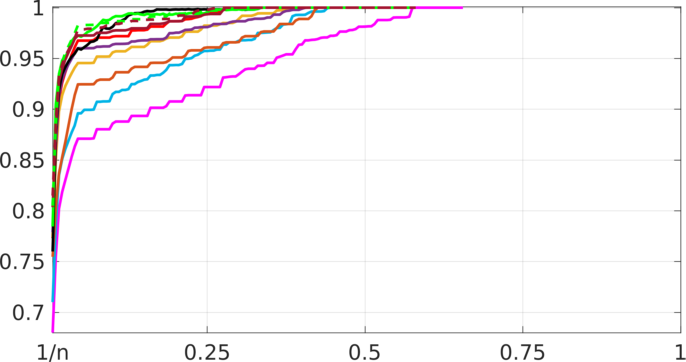}};
\draw (803,462) node  {\includegraphics[width=285pt,height=136.5pt]{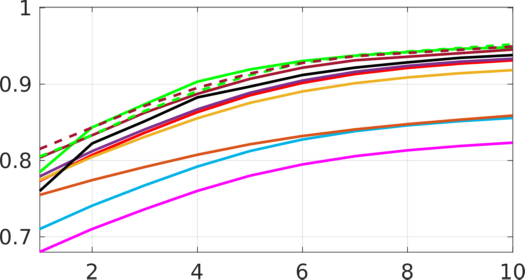}};

\draw (263,8) node [anchor=north west][inner sep=0.75pt]   [align=left] {\textbf{\textcolor[rgb]{0,0,0}{MARS}}};
\draw (749,306) node [anchor=north west][inner sep=0.75pt]   [align=left] {\textbf{\textcolor[rgb]{0,0,0}{P-DESTRE}}};
\draw (11.5,238.5) node [anchor=north west][inner sep=0.75pt]  [color={rgb, 255:red, 0; green, 0; blue, 0 }  ,opacity=1 ,rotate=-270] [align=left] {\textcolor[rgb]{0,0,0}{Identification Rate (Hit)}};
\draw (201,285) node [anchor=north west][inner sep=0.75pt]   [align=left] {\textcolor[rgb]{0,0,0}{Acc. Rank (Penetration)}};
\draw (511.5,526.5) node [anchor=north west][inner sep=0.75pt]  [color={rgb, 255:red, 0; green, 0; blue, 0 }  ,opacity=1 ,rotate=-270] [align=left] {\textcolor[rgb]{0,0,0}{Identification Rate (Hit)}};
\draw (716,588) node [anchor=north west][inner sep=0.75pt]   [align=left] {\textcolor[rgb]{0,0,0}{Acc. Rank (Penetration)}};
\draw (510.5,236.5) node [anchor=north west][inner sep=0.75pt]  [color={rgb, 255:red, 0; green, 0; blue, 0 }  ,opacity=1 ,rotate=-270] [align=left] {\textcolor[rgb]{0,0,0}{Identification Rate (Hit)}};
\draw (708,292) node [anchor=north west][inner sep=0.75pt]   [align=left] {\textcolor[rgb]{0,0,0}{Acc. Rank (Penetration)}};
\draw (755,13) node [anchor=north west][inner sep=0.75pt]   [align=left] {\textcolor[rgb]{0,0,0}{\textbf{{\small iLIDS-VID}}}};
\draw (210,583) node [anchor=north west][inner sep=0.75pt]   [align=left] {\textcolor[rgb]{0,0,0}{Acc. Rank (Penetration)}};
\draw (8.5,521.5) node [anchor=north west][inner sep=0.75pt]  [color={rgb, 255:red, 0; green, 0; blue, 0 }  ,opacity=1 ,rotate=-270] [align=left] {\textcolor[rgb]{0,0,0}{Identification Rate (Hit)}};
\draw (242,302) node [anchor=north west][inner sep=0.75pt]   [align=left] {\textbf{\textcolor[rgb]{0,0,0}{PRID2011}}};
\draw (211,610) node [anchor=north west][inner sep=0.75pt]   [align=left] {\textbf{\textcolor[rgb]{0.74,0.06,0.88}{Chung \etal \cite{chung2017two} }}};
\draw (209,631) node [anchor=north west][inner sep=0.75pt]   [align=left] {\textbf{\textcolor[rgb]{0.82,0.01,0.11}{Gao \& Nevatia \cite{gao2018revisiting}}}};
\draw (211,652) node [anchor=north west][inner sep=0.75pt]   [align=left] {\textbf{\textcolor[rgb]{0.29,0.56,0.89}{Rao \etal \cite{rao2018video}}}};
\draw (428,611) node [anchor=north west][inner sep=0.75pt]  [font=\normalsize,color={rgb, 255:red, 218; green, 117; blue, 11 }  ,opacity=1 ] [align=left] {\textbf{\textcolor[rgb]{0.8,0.38,0.07}{STAM   \cite{li2018diversity}}}};
\draw (429,632) node [anchor=north west][inner sep=0.75pt]  [color={rgb, 255:red, 218; green, 117; blue, 11 }  ,opacity=1 ] [align=left] {\textbf{\textcolor[rgb]{0.96,0.65,0.14}{TKP   \cite{gu2019temporal}}}};
\draw (428,655) node [anchor=north west][inner sep=0.75pt]  [color={rgb, 255:red, 218; green, 117; blue, 11 }  ,opacity=1 ] [align=left] {\textbf{\textcolor[rgb]{0.56,0.07,1}{STMPM  \cite{liu2019spatial}}}};
\draw (622,612) node [anchor=north west][inner sep=0.75pt]  [color={rgb, 255:red, 218; green, 117; blue, 11 }  ,opacity=1 ] [align=left] {\textbf{\textcolor[rgb]{0,0,0}{GLTR  \cite{li2019global} }}};
\draw (621,631) node [anchor=north west][inner sep=0.75pt]  [color={rgb, 255:red, 218; green, 117; blue, 11 }  ,opacity=1 ] [align=left] {\textbf{\textcolor[rgb]{0.51,0.91,0.08}{COSAM+Avg \cite{subramaniam2019co}}}};
\draw (623,653) node [anchor=north west][inner sep=0.75pt]  [color={rgb, 255:red, 218; green, 117; blue, 11 }  ,opacity=1 ] [align=left] {\textbf{\textcolor[rgb]{0.61,0.1,0.16}{NVAN+STE \cite{liu2019}}}};
\draw (807,614) node [anchor=north west][inner sep=0.75pt]  [color={rgb, 255:red, 218; green, 117; blue, 11 }  ,opacity=1 ] [align=left] {\textbf{\textcolor[rgb]{0.51,0.91,0.08}{COSAM+Pro}}};
\draw (811,646) node [anchor=north west][inner sep=0.75pt]  [color={rgb, 255:red, 218; green, 117; blue, 11 }  ,opacity=1 ] [align=left] {\textbf{\textcolor[rgb]{0.61,0.1,0.16}{NVAN+Pro}}};

\end{tikzpicture}
}
 \caption{ Comparison between the CMC curves
observed for the state-of-the-art methods in the four datasets considered. Zoomed-in regions inside each plot provide the top-i, $\forall i \in \{1\ldots 10\}$ values.}
\label{overall_g}
\end{figure*}

This section provides a comparison between the results obtained when using the symbolic representations proposed and the techniques considered to represent the state-of-the-art. Fig. \ref{overall_g} provides the CMC curves for all methods and datasets. Overall, it can be observed that the proposed symbolic representation tied to the NVAN method got the best results in all datasets, with exception to the iLIDS-VID set.

In particular, Table \ref{mars_overall} summarizes the performance attained in the MARS dataset.  When looking to the mAP measure, the baseline model NVAN  \cite{liu2019} and COSAM  \cite{subramaniam2019co} outperform the other methods. Also, the proposed temporal pooling method  increased the mAP values by  0.77\% in the COSAM   and 0.93\% in the NVAN  model, when compared to the original representations. In terms of rank-1, NVAN  and GLTR \cite{li2019global}  stood in top two positions, with the proposed temporal pooling method increasing the rank-1 accuracy by 2.11\% in COSAM  and 1.36\% in NVAN techniques. For higher (and less significant ranks) ranks, both COSAM and NVAN provided similar levels of performance.

A similar comparison regarding the iLIDS-VID set is provided in Table \ref{ilidsvid_overall}.  Here, the STMPM  \cite{liu2019spatial} and NVAN  \cite{liu2019} stood in the second and third best positions, respectively. Again, the proposed temporal model increased the baseline results by 3.51\% in COSAM \cite{subramaniam2019co} and 0.32 \% in NVAN for rank-1. For   higher ranks, NVAN showed superior performance than the GLTR.

Next, Table \ref{prid_overall} compares the results for the PRID2011 dataset.  For rank-1, GLTR \cite{li2019global} outperformed the other methods, with NVAN \cite{liu2019} standing as runner-up technique. The proposed temporal pooling method improved the effectiveness of COSAM by 1.78\% and of NVAN by 1.92\%. 

Finally, Table \ref{pdestre_overall} provides the performance values observed for UAV-based P-DESTRE dataset. Here, COSAM \cite{subramaniam2019co}  and NVAN \cite{liu2019} outperformed the other approaches in terms of mAP, rank-1 and rank-20. As in the previous sets, the proposed temporal method increased the baseline model by 1.07\% (COSAM) and 1.43\% (NVAN) in terms of the mAP values, and by 1.42\% (COSAM) and 2.32\% (NVAN) in terms of rank-1 results.

\begin{table}[h]
\centering
\caption{ \scriptsize{State-of-the-art comparison (MARS dataset)}}
\label{mars_overall}
\begin{tabular}{c|c|c|c}
\hline
 \scriptsize{\textbf{Method} }    &  \scriptsize{\textbf{mAP}}                           &  \scriptsize{\textbf{R1}}                            &  \scriptsize{\textbf{R20}}                           \\ \hline
 \scriptsize{Chung \etal \cite{chung2017two} }     & \scriptsize{65.70}\tiny{$\pm$} 3.24  &  \scriptsize{73.54} \tiny{$\pm$} 1.82 &  \scriptsize{86.18} \tiny{$\pm$} 1.17 \\ \hline
 \scriptsize{Gao \& Nevatia \cite{gao2018revisiting} }&  \scriptsize{73.50} \tiny{$\pm$} 2.74 &  \scriptsize{82.90} \tiny{$\pm$} 1.15 &  \scriptsize{96.52} \tiny{$\pm$} 1.08 \\ \hline
 \scriptsize{Rao \etal \cite{rao2018video} }    &  \scriptsize{70.15} \tiny{$\pm$} 3.17 &  \scriptsize{76.58} \tiny{$\pm$} 1.29 &  \scriptsize{89.24} \tiny{$\pm$} 1.27 \\ \hline
 \scriptsize{STAM   \cite{li2018diversity} }     &  \scriptsize{64.94} \tiny{$\pm$} 2.40 &  \scriptsize{81.75} \tiny{$\pm$} 1.96 &  \scriptsize{95.54} \tiny{$\pm$} 1.28 \\ \hline
 \scriptsize{TKP   \cite{gu2019temporal} }     &  \scriptsize{72.82} \tiny{$\pm$} 2.24 &  \scriptsize{82.93} \tiny{$\pm$} 1.83 &  \scriptsize{95.11} \tiny{$\pm$} 1.37 \\ \hline
 \scriptsize{STMPM  \cite{liu2019spatial}}       &  \scriptsize{71.36} \tiny{$\pm$} 2.36 &  \scriptsize{83.50} \tiny{$\pm$} 1.97 &  \scriptsize{96.58} \tiny{$\pm$} 1.26 \\ \hline
 \scriptsize{GLTR}  \cite{li2019global}      &  \scriptsize{77.74} \tiny{$\pm$} 1.07 &  \scriptsize{84.72} \tiny{$\pm$} 2.61 &  \scriptsize{95.79} \tiny{$\pm$} 2.34 \\ \hline
 \scriptsize{COSAM+Avg} \cite{subramaniam2019co} &  \scriptsize{78.35}\tiny{$\pm$} 1.66 &  \scriptsize{84.03}\tiny{$\pm$} 0.91 &  \scriptsize{96.98} \tiny{$\pm$} 0.98  \\ \hline
 \scriptsize{NVAN+STE }\cite{liu2019} &
   \scriptsize{81.13} \tiny{$\pm$} 1.57 &
   \scriptsize{85.94} \tiny{$\pm$} 0.94 &
   \scriptsize{97.26} \tiny{$\pm$} 0.97 \\ \hline
 \scriptsize{\textbf{COSAM+Pro}} &
  \textbf{\scriptsize{79.12} \tiny{$\pm$} 1.35} &
 \textbf{  \scriptsize{86.14} \tiny{$\pm$} 0.84} &
 \textbf{  \scriptsize{97.52} \tiny{$\pm$} 0.91}\\ \hline
 \scriptsize{\textbf{NVAN+Pro}} &
 \textbf{  \scriptsize{82.06} \tiny{$\pm$} 1.53} &
  \textbf{ \scriptsize{87.30} \tiny{$\pm$} 0.89} &
  \textbf{ \scriptsize{97.80} \tiny{$\pm$} 0.84} \\ \hline
\end{tabular}%
\end{table}

\begin{table}[h]
\centering
\caption{\scriptsize{State-of-the-art comparison (iLIDS-VID dataset)}}
\label{ilidsvid_overall}
\begin{tabular}{c|c|c}
\hline
 \scriptsize{\textbf{Method }}         &  \scriptsize{\textbf{R1  }}                          &  \scriptsize{\textbf{ R20 } }                         \\ \hline
 \scriptsize{Chung \etal \cite{chung2017two}}      &  \scriptsize{60.26} \tiny{$\pm$} 4.54   &  \scriptsize{96.10} \tiny{$\pm$} 1.30   \\ \hline
 \scriptsize{Gao \& Nevatia \cite{gao2018revisiting}} &  \scriptsize{57.58} \tiny{$\pm$} 4.59   &  \scriptsize{97.21} \tiny{$\pm$} 1.12   \\ \hline
 \scriptsize{Rao  \etal \cite{rao2018video} }      &  \scriptsize{65.68} \tiny{$\pm$} 3.54   &  \scriptsize{98.75} \tiny{$\pm$} 1.14   \\ \hline
 \scriptsize{STAM \cite{li2018diversity}}           &  \scriptsize{79.72} \tiny{$\pm$}  3.31   &  \scriptsize{95.50} \tiny{$\pm$}  1.51   \\ \hline
 \scriptsize{TKP  \cite{gu2019temporal}}             &  \scriptsize{66.11} \tiny{$\pm$}   3.84  &  \scriptsize{98.00} \tiny{$\pm$}  1.95   \\ \hline
 \scriptsize{STMPM \cite{liu2019spatial} }          &  \scriptsize{83.73} \tiny{$\pm$}  3.99   &  \scriptsize{99.16} \tiny{$\pm$}  1.11   \\ \hline
 \scriptsize{GLTR    \cite{li2019global} }        &  \scriptsize{85.60} \tiny{$\pm$}  3.53   &  \scriptsize{99.07} \tiny{$\pm$}  1.05   \\ \hline
 \scriptsize{COSAM  \cite{subramaniam2019co}  }         &  \scriptsize{78.74 } \tiny{$\pm$}   4.10 &  \scriptsize{97.50}  \tiny{$\pm$}   1.31 \\ \hline
 \scriptsize{NVAN   \cite{liu2019}  }         &  \scriptsize{81.71 }\tiny{$\pm$}  3.27   &  \scriptsize{99.21} \tiny{$\pm$}  0.99   \\ \hline
 \scriptsize{\textbf{COSAM \cite{subramaniam2019co} +Pro   }}   &  \scriptsize{\textbf{82.25} \tiny{$\pm$}  3.16}   &  \scriptsize{\textbf{98.93} \tiny{$\pm$}  1.27 }  \\ \hline
 \scriptsize{\textbf{NVAN \cite{liu2019} +Pro}}       & \scriptsize{ \textbf{82.03} \tiny{$\pm$}  3.54}  &  \scriptsize{\textbf{99.32} \tiny{$\pm$}  1.14}   \\ \hline
\end{tabular}%

\end{table}

\begin{table}[h]
\centering
\caption{\scriptsize{State-of-the-art comparison (PRID2011 dataset)}}
\label{prid_overall}
\begin{tabular}{c|c|c}
\hline
\scriptsize{\textbf{Method }}          & \scriptsize{\textbf{R1  } }                          & \scriptsize{\textbf{R20 }    }                       \\ \hline
\scriptsize{Chung \etal \cite{chung2017two}  }     & \scriptsize{77.40} \tiny{$\pm$} 3.61 & \scriptsize{98.91} \tiny{$\pm$} 2.34 \\ \hline
\scriptsize{Gao \& Nevatia \cite{gao2018revisiting}}  & \scriptsize{79.59} \tiny{$\pm$} 3.14 & \scriptsize{99.25}  \tiny{$\pm$} 2.74 \\ \hline
\scriptsize{Rao  \etal \cite{rao2018video} }      & \scriptsize{88.08} \tiny{$\pm$} 2.96 & \scriptsize{99.00} \tiny{$\pm$} 1.79 \\ \hline
\scriptsize{STAM \cite{li2018diversity}}          & \scriptsize{93.08} \tiny{$\pm$} 2.51 & \scriptsize{98.41} \tiny{$\pm$} 0.34 \\ \hline
\scriptsize{TKP  \cite{gu2019temporal}}       & \scriptsize{81.63}  \tiny{$\pm$} 3.83 & \scriptsize{98.56}  \tiny{$\pm$} 1.04                 \\ \hline
\scriptsize{STMPM \cite{liu2019spatial}}            & \scriptsize{91.32} \tiny{$\pm$} 2.11 & \scriptsize{99.09} \tiny{$\pm$} 0.64 \\ \hline
\scriptsize{GLTR    \cite{li2019global} }           & \scriptsize{94.96} \tiny{$\pm$} 0.69 & \scriptsize{100} \tiny{$\pm$} 0   \\ \hline
\scriptsize{COSAM+Avg \cite{subramaniam2019co}}  &
  \scriptsize{84.21} \tiny{$\pm$} 3.07 &
  \scriptsize{99.51 } \tiny{$\pm$} 0.54  \\ \hline
\scriptsize{NVAN+STE \cite{liu2019}}  &
  \scriptsize{93.32}  \tiny{$\pm$} 2.54 &
  \scriptsize{100 } \tiny{$\pm$} 0  \\ \hline
\scriptsize{\textbf{COSAM+Pro}}  &
 \scriptsize{\textbf{ 85.99}  \tiny{$\pm$} 2.98} &
  \scriptsize{\textbf{99.86}  \tiny{$\pm$} 0.48}  \\ \hline
\scriptsize{\textbf{NVAN+Pro}}  &
 \scriptsize{ \textbf{95.24}  \tiny{$\pm$} 1.24} &
 \scriptsize{\textbf{ 100 } \tiny{$\pm$} 0}  \\ \hline
\end{tabular}%
\end{table}

\begin{table}[h]
\centering
\caption{\scriptsize{State-of-the-art comparison (P-DESTRE dataset}}
\label{pdestre_overall}
\begin{tabular}{c|c|c|c}
\hline
\scriptsize{\textbf{Method}  }   & \scriptsize{\textbf{mAP}    }                    & \scriptsize{\textbf{R1}}                            &\scriptsize{ \textbf{R20}    }                       \\ \hline
\scriptsize{Chung \etal \cite{chung2017two} }   & \scriptsize{67.77} \tiny{$\pm$} 11.62 & \scriptsize{68.01} \tiny{$\pm$} 13.76 & \scriptsize{86.52} \tiny{$\pm$} 3.47 \\ \hline
\scriptsize{Gao \&  Nevatia \cite{gao2018revisiting}}  & \scriptsize{75.57} \tiny{$\pm$} 10.69 & \scriptsize{77.30} \tiny{$\pm$} 12.65 & \scriptsize{96.24} \tiny{$\pm$} 4.16 \\ \hline
\scriptsize{Rao \etal \cite{rao2018video} }      & \scriptsize{72.22} \tiny{$\pm$} 11.07 & \scriptsize{71.01} \tiny{$\pm$} 12.71 & \scriptsize{88.56} \tiny{$\pm$} 4.13 \\ \hline
\scriptsize{STAM  \cite{li2018diversity} }    & \scriptsize{67.01} \tiny{$\pm$} 11.73 & \scriptsize{75.48} \tiny{$\pm$} 11.41 & \scriptsize{91.69} \tiny{$\pm$} 3.79 \\ \hline
\scriptsize{TKP  \cite{gu2019temporal} }      & \scriptsize{74.89} \tiny{$\pm$} 11.69 & \scriptsize{77.42} \tiny{$\pm$} 11.62 & \scriptsize{94.21} \tiny{$\pm$} 4.15 \\ \hline
\scriptsize{STMPM  \cite{liu2019spatial}}          & \scriptsize{73.43 }\tiny{$\pm$} 12.08 & \scriptsize{77.90 }\tiny{$\pm$} 13.72 & \scriptsize{95.61} \tiny{$\pm$} 4.71 \\ \hline
\scriptsize{GLTR  \cite{li2019global} }     & \scriptsize{77.68 }\tiny{$\pm$} 9.46  & \scriptsize{75.96} \tiny{$\pm$} 11.77 & \scriptsize{95.48} \tiny{$\pm$} 3.17 \\ \hline
\scriptsize{COSAM+Avg \cite{subramaniam2019co}} &
  \scriptsize{80.42} \tiny{$\pm$} 9.91 &
  \scriptsize{79.14 }\tiny{$\pm$} 12.43 &
  \scriptsize{97.10}\tiny{$\pm$} 1.85 \\ \hline
\scriptsize{NVAN+STE \cite{liu2019}} &
  \scriptsize{82.78} \tiny{$\pm$} 10.35 &
  \scriptsize{80.36} \tiny{$\pm$} 12.38 &
  \scriptsize{97.10} \tiny{$\pm$} 1.93 \\ \hline
\scriptsize{\textbf{COSAM+Pro}} &
 \scriptsize{ \textbf{81.49} \tiny{$\pm$} 9.64} &
\scriptsize{ \textbf{ 80.56} \tiny{$\pm$} 11.72} &
 \scriptsize{ \textbf{98.54} \tiny{$\pm$} 1.78} \\ \hline
\scriptsize{\textbf{NVAN+Pro}} &
\scriptsize{  \textbf{84.21} \tiny{$\pm$} 10.37} &
 \scriptsize{ \textbf{82.68 }\tiny{$\pm$} 11.51} &
\scriptsize{ \textbf{ 98.30} \tiny{$\pm$} 1.87} \\ \hline
\end{tabular}%
\end{table}

Note that, among the state-of-the-art methods, COSAM \cite{subramaniam2019co}  and TKP  \cite{gu2019temporal} use \textit{avg pooling} method for feature aggregation, while all other techniques either use attention or 3D-CNN models to obtain the video level representations. GLTR method showed a competitive performance in all datasets, using short- and long-term temporal cues to learn discriminant video representations. Here, the short term cues appear to have helped to distinguish among adjacent frames where the long terms cues help to handle occlusion. 

On the MARS, PRID2011, and P-DESTRE datasets, the NVAN method provided the best levels of performance, in particular when used along to the proposed temporal polling method. In the iLIDS-VID dataset, GLTR outperformed any other method, which we justified by the large fraction of partially occluded sample in this set, and the way that this algorithm handles occlusions by short- and long-term features. Among all the datasets considered, PRID2011 is the one of smallest dimensions, and we observed that levels of performance were close to saturation in some cases. In opposition, as MARS and P-DESTRE are relatively larger than the other datasets, they can also be considered more challenging, in terms of the data degradation factors yielding from the data acquisition protocols used in such sets. Interestingly, the P-DESTRE set provided better results than MARS, in particular for  high ranking values, probably as a consequence of a larger tracklet length and less degraded samples in P-DESTRE than in MARS. 

\begin{figure*}
 \centering
 \resizebox*{!}{\textheight}{%
\tikzset{every picture/.style={line width=0.75pt}} 
\begin{tikzpicture}[x=0.75pt,y=0.75pt,yscale=-0.95,xscale=1]

\draw (61.25,95.5) node  {\includegraphics[width=30.37pt,height=90.75pt]{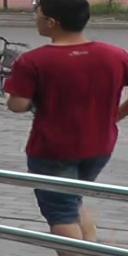}};
\draw (103.25,95.5) node  {\includegraphics[width=31.12pt,height=90.75pt]{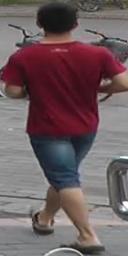}};
\draw (145.75,95.5) node  {\includegraphics[width=31.12pt,height=90.75pt]{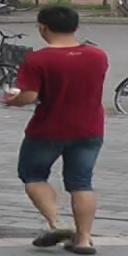}};
\draw (188.25,95.5) node  {\includegraphics[width=31.13pt,height=90.75pt]{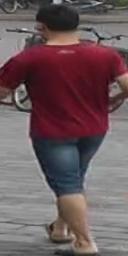}};
\draw (255.25,97) node  {\includegraphics[width=30.38pt,height=91.5pt]{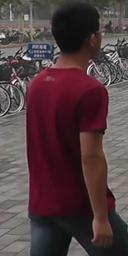}};
\draw (297.5,97) node  {\includegraphics[width=31.5pt,height=91.5pt]{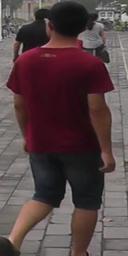}};
\draw (339.5,97) node  {\includegraphics[width=30pt,height=91.5pt]{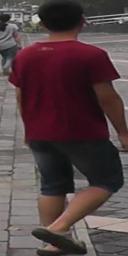}};
\draw (382,97) node  {\includegraphics[width=32.25pt,height=91.5pt]{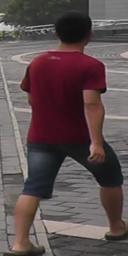}};
\draw (61.5,247.5) node  {\includegraphics[width=31.5pt,height=92.25pt]{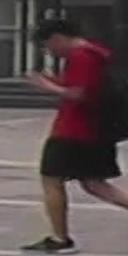}};
\draw (104.5,247.5) node  {\includegraphics[width=31.5pt,height=92.25pt]{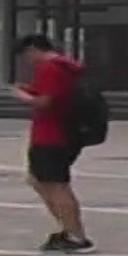}};
\draw (147.5,247) node  {\includegraphics[width=31.5pt,height=91.5pt]{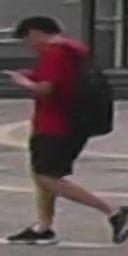}};
\draw (190,247) node  {\includegraphics[width=30.75pt,height=91.5pt]{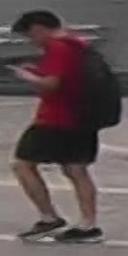}};
\draw (255.5,245.5) node  {\includegraphics[width=30pt,height=90.75pt]{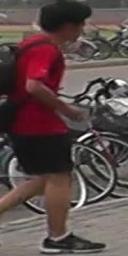}};
\draw (298.5,245.5) node  {\includegraphics[width=31.5pt,height=90.75pt]{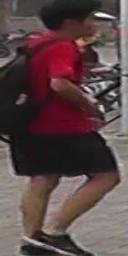}};
\draw (341,245.5) node  {\includegraphics[width=30.75pt,height=90.75pt]{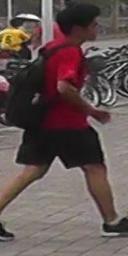}};
\draw (385,245.5) node  {\includegraphics[width=32.25pt,height=90.75pt]{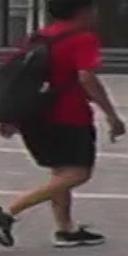}};
\draw [line width=2.25]  [dash pattern={on 2.53pt off 3.02pt}]  (420.5,18) -- (427.5,318) -- (427.5,318) -- (434.5,1308) ;
\draw (464.25,97.5) node  {\includegraphics[width=30.38pt,height=92.25pt]{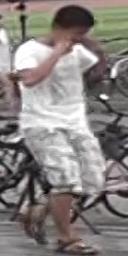}};
\draw (506.25,97.5) node  {\includegraphics[width=31.13pt,height=92.25pt]{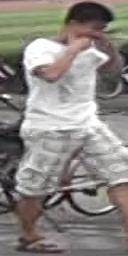}};
\draw (548.25,97.5) node  {\includegraphics[width=30.38pt,height=92.25pt]{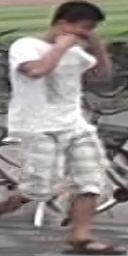}};
\draw (590.75,97.5) node  {\includegraphics[width=31.88pt,height=92.25pt]{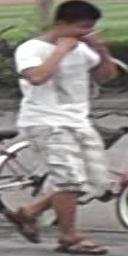}};
\draw (661.25,97.5) node  {\includegraphics[width=33.38pt,height=92.25pt]{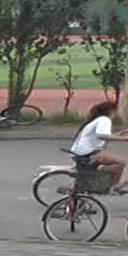}};
\draw (708.75,97.5) node  {\includegraphics[width=34.88pt,height=92.25pt]{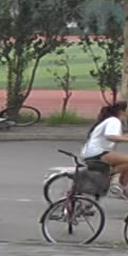}};
\draw (756.75,97.5) node  {\includegraphics[width=35.63pt,height=92.25pt]{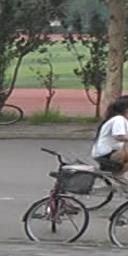}};
\draw (805.25,97.5) node  {\includegraphics[width=35.63pt,height=92.25pt]{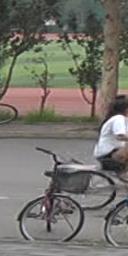}};
\draw (467,245.5) node  {\includegraphics[width=32.25pt,height=90.75pt]{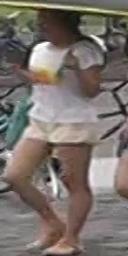}};
\draw (510.75,245.5) node  {\includegraphics[width=31.13pt,height=90.75pt]{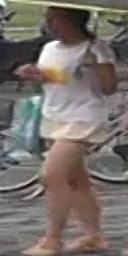}};
\draw (551.75,245) node  {\includegraphics[width=28.13pt,height=90pt]{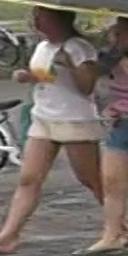}};
\draw (593.5,245.5) node  {\includegraphics[width=33pt,height=90.75pt]{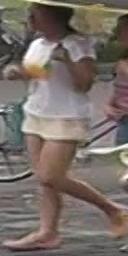}};
\draw (662,245) node  {\includegraphics[width=33.75pt,height=90pt]{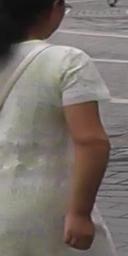}};
\draw (709.5,245) node  {\includegraphics[width=36pt,height=90pt]{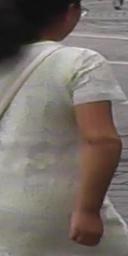}};
\draw (758.5,245) node  {\includegraphics[width=34.5pt,height=90pt]{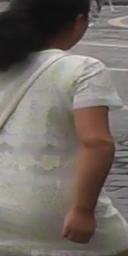}};
\draw (807,245) node  {\includegraphics[width=35.25pt,height=90pt]{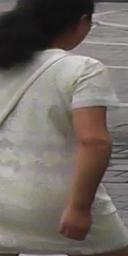}};
\draw (60.5,413.5) node  {\includegraphics[width=30pt,height=93.75pt]{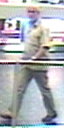}};
\draw (104,412.5) node  {\includegraphics[width=32.25pt,height=92.25pt]{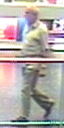}};
\draw (148,411.5) node  {\includegraphics[width=32.25pt,height=92.25pt]{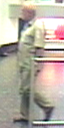}};
\draw (192,411) node  {\includegraphics[width=29.25pt,height=91.5pt]{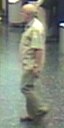}};
\draw (256,409.5) node  {\includegraphics[width=30.75pt,height=90.75pt]{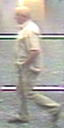}};
\draw (300.5,409) node  {\includegraphics[width=31.5pt,height=90pt]{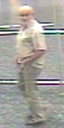}};
\draw (344.5,409) node  {\includegraphics[width=31.5pt,height=90pt]{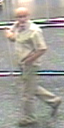}};
\draw (388,409) node  {\includegraphics[width=30.75pt,height=90pt]{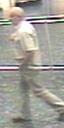}};
\draw (60.5,566) node  {\includegraphics[width=30pt,height=93pt]{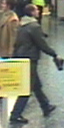}};
\draw (105,565.5) node  {\includegraphics[width=33.75pt,height=93.75pt]{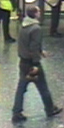}};
\draw (150,566) node  {\includegraphics[width=30.75pt,height=94.5pt]{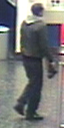}};
\draw (193.25,565.5) node  {\includegraphics[width=29.63pt,height=93.75pt]{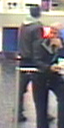}};
\draw (255.5,565.5) node  {\includegraphics[width=32.25pt,height=93.75pt]{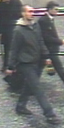}};
\draw (300.5,564.5) node  {\includegraphics[width=31.5pt,height=93.75pt]{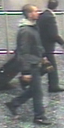}};
\draw (344.5,564) node  {\includegraphics[width=30pt,height=93pt]{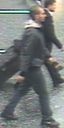}};
\draw (388.5,564) node  {\includegraphics[width=31.5pt,height=93pt]{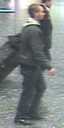}};
\draw (470,409) node  {\includegraphics[width=32.25pt,height=90pt]{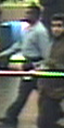}};
\draw (513.5,408.5) node  {\includegraphics[width=30pt,height=90.75pt]{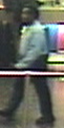}};
\draw (554,408.5) node  {\includegraphics[width=27.75pt,height=90.75pt]{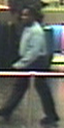}};
\draw (595.5,408.5) node  {\includegraphics[width=31.5pt,height=90.75pt]{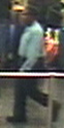}};
\draw (662.5,408) node  {\includegraphics[width=34.5pt,height=90pt]{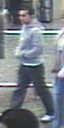}};
\draw (712.75,408) node  {\includegraphics[width=37.13pt,height=90pt]{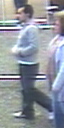}};
\draw (763.5,408) node  {\includegraphics[width=34.5pt,height=90pt]{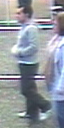}};
\draw (810,408) node  {\includegraphics[width=30.75pt,height=90pt]{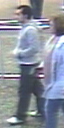}};
\draw (471,564) node  {\includegraphics[width=30.75pt,height=93pt]{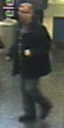}};
\draw (515,564) node  {\includegraphics[width=32.25pt,height=93pt]{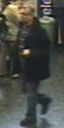}};
\draw (557,564) node  {\includegraphics[width=27.75pt,height=93pt]{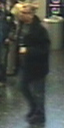}};
\draw (597.5,563.5) node  {\includegraphics[width=30pt,height=92.25pt]{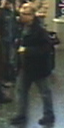}};
\draw (662.5,564.5) node  {\includegraphics[width=34.5pt,height=92.25pt]{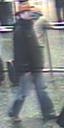}};
\draw (714,565) node  {\includegraphics[width=38.25pt,height=93pt]{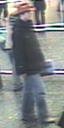}};
\draw (765.5,564.5) node  {\includegraphics[width=33pt,height=92.25pt]{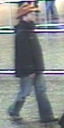}};
\draw (810.5,564) node  {\includegraphics[width=30pt,height=91.5pt]{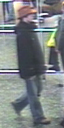}};
\draw (59.5,732) node  {\includegraphics[width=30pt,height=93pt]{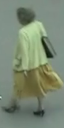}};
\draw (103.5,732) node  {\includegraphics[width=33pt,height=93pt]{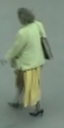}};
\draw (148.5,732) node  {\includegraphics[width=31.5pt,height=93pt]{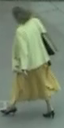}};
\draw (193,732) node  {\includegraphics[width=30.75pt,height=93pt]{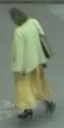}};
\draw (256.75,731.5) node  {\includegraphics[width=33.38pt,height=92.25pt]{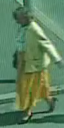}};
\draw (302.75,731.5) node  {\includegraphics[width=32.63pt,height=92.25pt]{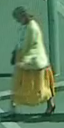}};
\draw (347,731.5) node  {\includegraphics[width=29.25pt,height=92.25pt]{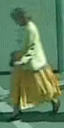}};
\draw (390,731.5) node  {\includegraphics[width=32.25pt,height=92.25pt]{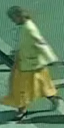}};
\draw (58.5,887) node  {\includegraphics[width=31.5pt,height=93pt]{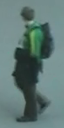}};
\draw (104,887) node  {\includegraphics[width=32.25pt,height=93pt]{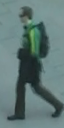}};
\draw (147.5,887) node  {\includegraphics[width=30pt,height=93pt]{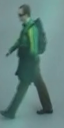}};
\draw (192,887) node  {\includegraphics[width=33.75pt,height=93pt]{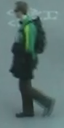}};
\draw (256.5,887.5) node  {\includegraphics[width=33pt,height=92.25pt]{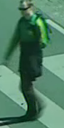}};
\draw (303,887.5) node  {\includegraphics[width=33.75pt,height=92.25pt]{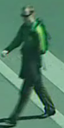}};
\draw (349,887.5) node  {\includegraphics[width=30.75pt,height=92.25pt]{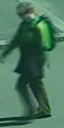}};
\draw (393.5,887.5) node  {\includegraphics[width=33pt,height=92.25pt]{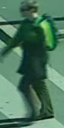}};
\draw (473.5,731.5) node  {\includegraphics[width=30pt,height=92.25pt]{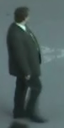}};
\draw (517,731) node  {\includegraphics[width=33.75pt,height=91.5pt]{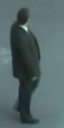}};
\draw (559.5,731) node  {\includegraphics[width=27pt,height=91.5pt]{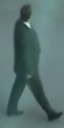}};
\draw (598,731) node  {\includegraphics[width=27.75pt,height=91.5pt]{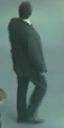}};
\draw (663,730) node  {\includegraphics[width=36pt,height=92.25pt]{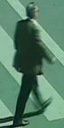}};

\draw (713.5,730.5) node  {\includegraphics[width=36pt,height=92.25pt]{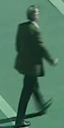}};
\draw (764,731) node  {\includegraphics[width=33.75pt,height=93pt]{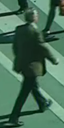}};
\draw (811,731.5) node  {\includegraphics[width=30.75pt,height=93.75pt]{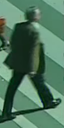}};
\draw (473.5,887.5) node  {\includegraphics[width=30pt,height=92.25pt]{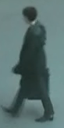}};
\draw (517,887.5) node  {\includegraphics[width=30.75pt,height=92.25pt]{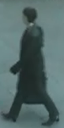}};
\draw (559,887) node  {\includegraphics[width=27.75pt,height=91.5pt]{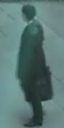}};
\draw (599.5,887) node  {\includegraphics[width=30pt,height=91.5pt]{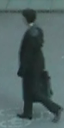}};
\draw (662,886) node  {\includegraphics[width=35.25pt,height=91.5pt]{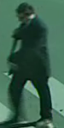}};
\draw (714,886.5) node  {\includegraphics[width=38.25pt,height=92.25pt]{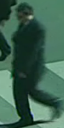}};
\draw (766,887) node  {\includegraphics[width=35.25pt,height=93pt]{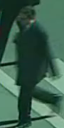}};
\draw (812,886.5) node  {\includegraphics[width=30.75pt,height=92.25pt]{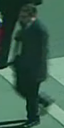}};
\draw (60.75,1053) node  {\includegraphics[width=31.13pt,height=94.5pt]{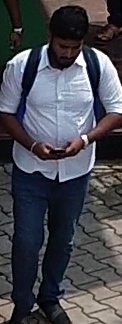}};
\draw (104.5,1053) node  {\includegraphics[width=31.5pt,height=94.5pt]{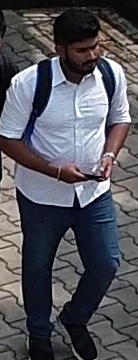}};
\draw (150,1053) node  {\includegraphics[width=32.25pt,height=94.5pt]{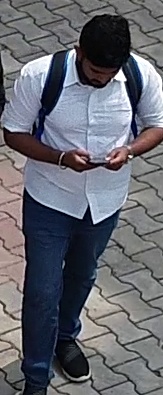}};
\draw (193.5,1053.5) node  {\includegraphics[width=30pt,height=95.25pt]{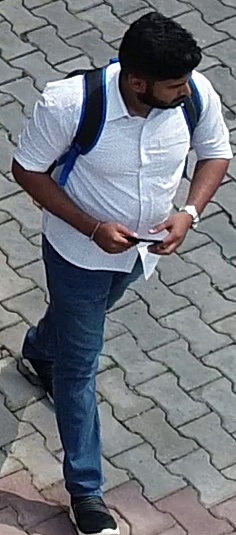}};
\draw (257.5,1053.5) node  {\includegraphics[width=33pt,height=95.25pt]{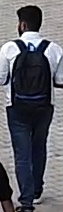}};
\draw (305.5,1053.5) node  {\includegraphics[width=34.5pt,height=95.25pt]{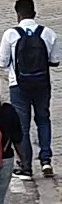}};
\draw (349.5,1053.5) node  {\includegraphics[width=28.5pt,height=95.25pt]{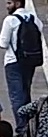}};
\draw (392.25,1053.5) node  {\includegraphics[width=29.63pt,height=95.25pt]{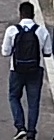}};
\draw (61,1215.5) node  {\includegraphics[width=32.25pt,height=95.25pt]{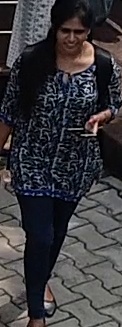}};
\draw (106,1215.5) node  {\includegraphics[width=32.25pt,height=95.25pt]{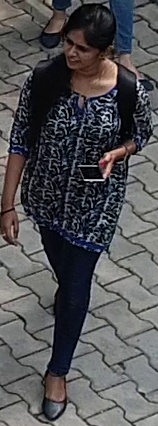}};
\draw (152.5,1216) node  {\includegraphics[width=33pt,height=96pt]{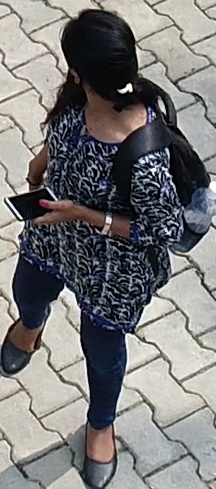}};
\draw (196,1216) node  {\includegraphics[width=27.75pt,height=96pt]{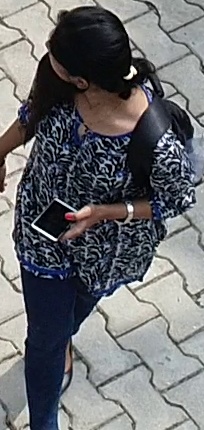}};
\draw (257,1215) node  {\includegraphics[width=38.25pt,height=96pt]{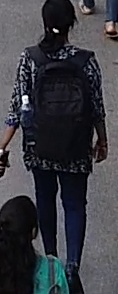}};
\draw (309,1214.5) node  {\includegraphics[width=36.75pt,height=96.75pt]{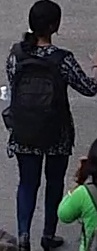}};
\draw (355.5,1214.5) node  {\includegraphics[width=30pt,height=96.75pt]{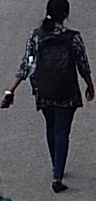}};
\draw (396,1214.5) node  {\includegraphics[width=27.75pt,height=96.75pt]{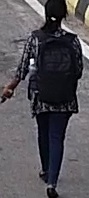}};
\draw (472,1054.5) node  {\includegraphics[width=32.25pt,height=95.25pt]{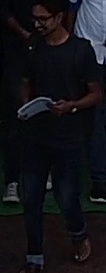}};
\draw (518.5,1054) node  {\includegraphics[width=33pt,height=96pt]{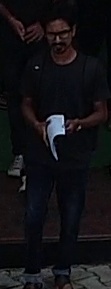}};
\draw (563.5,1054) node  {\includegraphics[width=31.5pt,height=96pt]{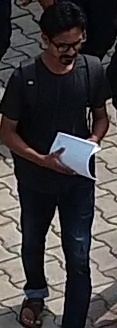}};
\draw (605.5,1054) node  {\includegraphics[width=28.5pt,height=96pt]{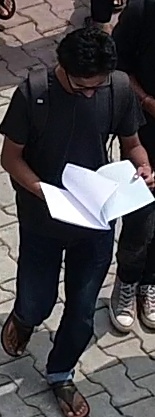}};
\draw (666.5,1054.5) node  {\includegraphics[width=36pt,height=96.75pt]{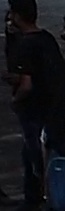}};
\draw (717.5,1054.5) node  {\includegraphics[width=37.5pt,height=96.75pt]{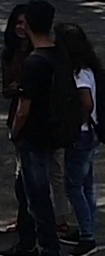}};
\draw (765.5,1054.5) node  {\includegraphics[width=31.5pt,height=96.75pt]{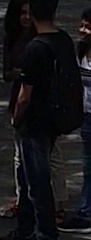}};
\draw (811.5,1054.5) node  {\includegraphics[width=33pt,height=96.75pt]{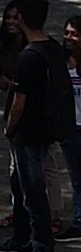}};
\draw (474,1215.5) node  {\includegraphics[width=33.75pt,height=98.25pt]{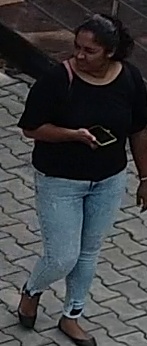}};
\draw (520.5,1215.5) node  {\includegraphics[width=34.5pt,height=98.25pt]{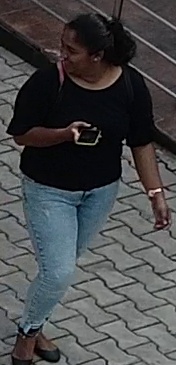}};
\draw (566,1215.5) node  {\includegraphics[width=30.75pt,height=98.25pt]{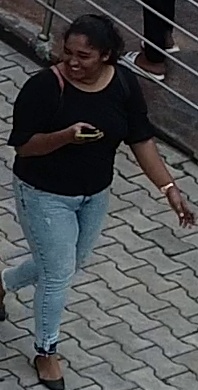}};
\draw (608.5,1215.5) node  {\includegraphics[width=30pt,height=98.25pt]{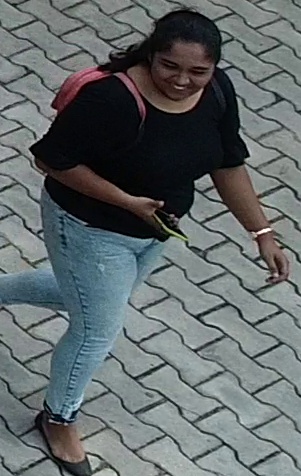}};
\draw (669.25,1215.5) node  {\includegraphics[width=33.38pt,height=96.75pt]{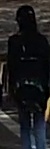}};
\draw (718.5,1215.5) node  {\includegraphics[width=36pt,height=96.75pt]{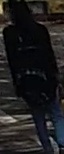}};
\draw (768.5,1214.5) node  {\includegraphics[width=34.5pt,height=96.75pt]{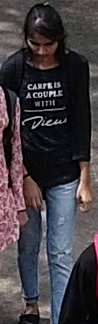}};
\draw (815.75,1214) node  {\includegraphics[width=30.38pt,height=96pt]{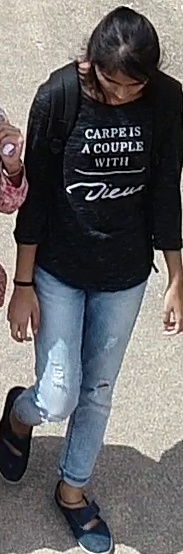}};

\draw (102,15) node [anchor=north west][inner sep=0.75pt]   [align=left] {\textbf{Query}};
\draw (292,15) node [anchor=north west][inner sep=0.75pt]   [align=left] {\textbf{Genuine}};
\draw (237,161) node [anchor=north west][inner sep=0.75pt]   [align=left] {\textbf{cdf$_{\text{avg}}$ 0.13, cdf$_{\text{pro}}$ 0.02}};
\draw (506,14) node [anchor=north west][inner sep=0.75pt]   [align=left] {\textbf{Query}};
\draw (704,15) node [anchor=north west][inner sep=0.75pt]   [align=left] {\textbf{Imposter}};
\draw (654.5,160.5) node [anchor=north west][inner sep=0.75pt]   [align=left] {\textbf{cdf$_{\text{avg}}$ 0.28, cdf$_{\text{pro}}$ 0.39}};
\draw (237.5,309) node [anchor=north west][inner sep=0.75pt]   [align=left] {\textbf{cdf$_{\text{avg}}$ 0.19, cdf$_{\text{pro}}$ 0.07}};
\draw (657.5,308.5) node [anchor=north west][inner sep=0.75pt]   [align=left] {\textbf{cdf$_{\text{avg}}$ 0.18, cdf$_{\text{pro}}$ 0.25}};
\draw (14.5,196.5) node [anchor=north west][inner sep=0.75pt]  [rotate=-270] [align=left] {\textbf{MARS}};
\draw (14.5,529.5) node [anchor=north west][inner sep=0.75pt]  [rotate=-270] [align=left] {\textbf{iLIDS-VID}};
\draw (237.5,473) node [anchor=north west][inner sep=0.75pt]   [align=left] {\textbf{cdf$_{\text{avg}}$ 0.21, cdf$_{\text{pro}}$ 0.04}};
\draw (240.5,629) node [anchor=north west][inner sep=0.75pt]   [align=left] {\textbf{cdf$_{\text{avg}}$ 0.16, cdf$_{\text{pro}}$ 0.03}};
\draw (660.5,470) node [anchor=north west][inner sep=0.75pt]   [align=left] {\textbf{cdf$_{\text{avg}}$ 0.14, cdf$_{\text{pro}}$ 0.25}};
\draw (661.5,628) node [anchor=north west][inner sep=0.75pt]   [align=left] {\textbf{cdf$_{\text{avg}}$ 0.11, cdf$_{\text{pro}}$ 0.22}};
\draw (242.5,799) node [anchor=north west][inner sep=0.75pt]   [align=left] {\textbf{cdf$_{\text{avg}}$ 0.10, cdf$_{\text{pro}}$ 0.03}};
\draw (244.5,954) node [anchor=north west][inner sep=0.75pt]   [align=left] {\textbf{cdf$_{\text{avg}}$ 0.15, cdf$_{\text{pro}}$ 0.01}};
\draw (17.5,844.5) node [anchor=north west][inner sep=0.75pt]  [rotate=-270] [align=left] {\textbf{PRID2011}};
\draw (660.5,796) node [anchor=north west][inner sep=0.75pt]   [align=left] {\textbf{cdf$_{\text{avg}}$ 0.10, cdf$_{\text{pro}}$ 0.20}};
\draw (660.5,952) node [anchor=north west][inner sep=0.75pt]   [align=left] {\textbf{cdf$_{\text{avg}}$ 0.16, cdf$_{\text{pro}}$ 0.29}};
\draw (245.5,1122) node [anchor=north west][inner sep=0.75pt]   [align=left] {\textbf{cdf$_{\text{avg}}$ 0.32, cdf$_{\text{pro}}$ 0.08}};
\draw (246.5,1286) node [anchor=north west][inner sep=0.75pt]   [align=left] {\textbf{cdf$_{\text{avg}}$ 0.38, cdf$_{\text{pro}}$ 0.14}};
\draw (17.5,1173.5) node [anchor=north west][inner sep=0.75pt]  [rotate=-270] [align=left] {\textbf{P-DESTRE}};
\draw (672.5,1287) node [anchor=north west][inner sep=0.75pt]   [align=left] {\textbf{cdf$_{\text{avg}}$ 0.12, cdf$_{\text{pro}}$ 0.36}};
\draw (665.5,1124) node [anchor=north west][inner sep=0.75pt]   [align=left] {\textbf{cdf$_{\text{avg}}$ 0.09cdf$_{\text{pro}}$ 0.33}};
\end{tikzpicture}
}
\caption{Examples of the sequences where the proposed method maximally increased performance, with respect to the traditional average pooling strategy. cdf$_\text{avg}$ and cdf$_\text{pro}$ denote the CDF values obtained by the SE-ResNet50+avg and SE-ResNet50+Pro models.}
\label{qualitative}
\end{figure*}
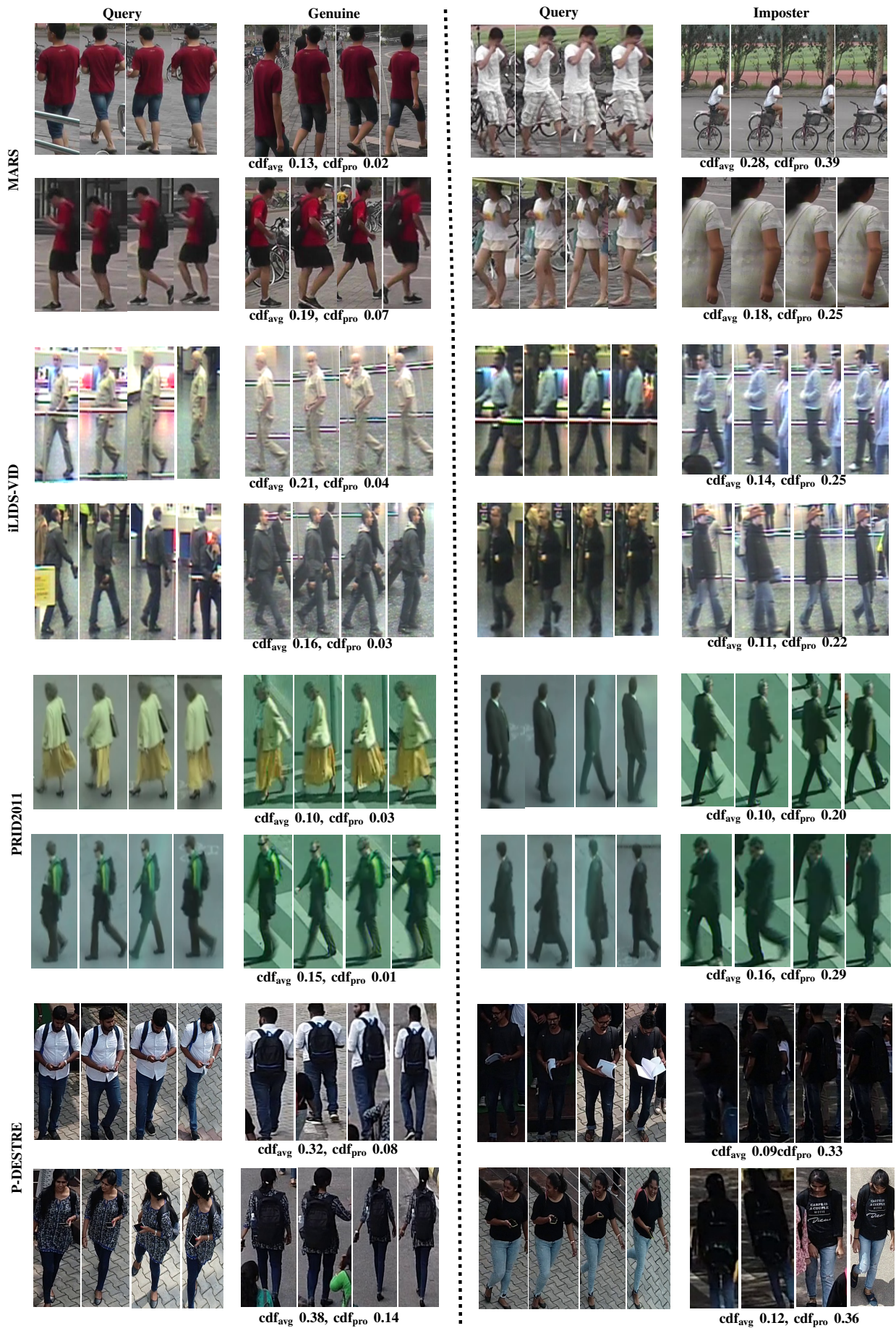

As an attempt to perceive the kind of samples where the proposed method most improves the results, Fig. \ref{qualitative} provides some examples of the optimal cases, i.e., those where, either in \emph{genuine} or \emph{impostors} comparisons,  the maximum absolute difference between the matching scores observed for the traditional pooling strategy(SE-ResNet50+avg) and for the method proposed in this paper (SE-ResNet50+pro methods). To do that, we estimated the continuous distribution function (cdf) of the genuine+impostor scores obtained for each data set, and then selected the pairs where the differences between the cdf$_{\text{avg}}$/cdf$_{\text{pro}}$ were maximal (note that for genuine comparisons \emph{better} scores correspond to smaller cdf values, with the opposite happening for impostor pairs). Overall, we observed that the maximum benefits of the proposed method occur in case of severe occlusions and for high heterogeneous sequences, in terms of the frames that compose them (e.g., with notorious differences in pose, lighting or background properties). 

Considering all the results given above, we concluded about the usability of the proposed symbolic representations for video based re-id. The experiments carried out point for:

\begin{itemize}
    \item Both the spatial and temporal features play orthogonal roles in the maximizing the discriminability between the different instances.  This is particularly evident in case of data of poor quality, where there is a reduced separability between the different classes (IDs).
    \item The backbone architecture plays an essential role in the effectiveness of the extracted feature sets, and determines strongly the final performance that can be attained in this problem. 
    \item Overall, the spatial attention-based models tend to perform better, when compared to other spatial feature extraction models.
    \item When comparing to the baselines, the proposed temporal pooling method is able to capture in a particularly effective way the variations among adjacent frames of highly heterogeneous tracklets, i.e., where the individual frames that compose the sequence have notoriously different appearance.
    \item Considering its  simple design, the proposed temporal pooling method is easily coupled to other spatial feature extractors and outperforms the baseline temporal cue learning models, either based in temporal attention and 3D-CNN models. 
\end{itemize}




 \section{Conclusions}
 \label{sec:Conclusions}

Motivated by the amount of discriminating information that is lost when using either the $avg$, $max$ or related pooling functions in the re-id deep-learning models, this paper we proposes a new temporal cue learning method based on symbolic representations that is able to faithfully fuse temporal and spatial cues to obtain effective feature representations. In our solution, the frame-level features are converted into their symbolic representation by fitting a distributional function for each feature. Each of these density functions captures the variability of a feature inside the corresponding tracklet, and can be understood as expressing the feature variability in a sequence. The idea is that symbolic representations keep more discriminating information of the original data than the classical pooling strategies, which provides the rationale for the observed  improvements in performance.  
Our experimental analysis was carried in four well known video-based re-id datasets. We compared our performance to the attained by nine techniques considered to represent the state-of-the-art. Overall, the observed results point for the superiority of the proposed method, in practically all datasets and performance measures, in particular when the images inside each sequence (tracklet) have large heterogeneity.

\section*{Acknowledgements}

This work is funded by FCT/MEC through national funds and  co-funded by FEDER - PT2020 partnership agreement under the projects UID/EEA/50008/2019, POCI-01-0247-FEDER-033395 and C4: Cloud Computing Competence Centre.

\bibliographystyle{ieee}
\bibliography{Main.bib}

\end{document}